\documentclass[journal]{IEEEtran}

\usepackage[utf8]{inputenc} 
\usepackage[T1]{fontenc}    
\usepackage{hyperref}       
\usepackage{url}            
\usepackage{booktabs}       
\usepackage{amsfonts}       
\usepackage{nicefrac}       
\usepackage{microtype}      
\usepackage{color}
\usepackage{algorithm,algpseudocode}
\usepackage{caption}
\usepackage{mwe}
\usepackage{lipsum}

\usepackage{graphicx}
\usepackage{amsmath}
\usepackage{float}
\usepackage{stfloats}
\usepackage{multirow}
\usepackage{amssymb}
\usepackage{amssymb}
\usepackage{graphicx}
\usepackage{pifont}
\newcommand{\cmark}{\ding{51}}%
\newcommand{\xmark}{\ding{55}}%
\usepackage{color}
\usepackage{cite}
\usepackage[hang,flushmargin]{footmisc} 
\usepackage{booktabs}
\usepackage[normalem]{ulem}
\newcommand{\revision}[1]{\textcolor{black}{#1}}
\newcommand{\revisions}[1]{\textcolor{black}{#1}}
\usepackage{soul}
\usepackage{float}
\soulregister{\cite}7
\soulregister{\ref}7
\soulregister\eqref7

\hypersetup{
    colorlinks=true,
    linkcolor=blue,
    filecolor=blue,
    urlcolor=blue,
    citecolor=blue
}

\ifCLASSINFOpdf
\else
\fi
\begin{document}
\bstctlcite{IEEEexample:BSTcontrol}

\title{Bayesian Neural Network Language Modeling for Speech Recognition }

\author{Boyang Xue,~\IEEEmembership
        Shoukang Hu,~\IEEEmembership
        Junhao Xu,~\IEEEmembership
        Mengzhe Geng,~\IEEEmembership\\
        Xunying Liu,~\IEEEmembership{Member,~IEEE,}
        Helen Meng,~\IEEEmembership{Fellow,~IEEE}
\thanks{
~~Boyang Xue, Shoukang Hu, Junhao Xu, Mengzhe Geng, Xunying Liu and Helen Meng are with the Department of System Engineering and Engineering Management, the Chinese University of Hong Kong, Hong Kong, China (email: byxue@se.cuhk.edu.hk; skhu@se.cuhk.edu.hk; jhxu@se.cuhk.edu.hk; mzgeng@se.cuhk.edu.hk; xyliu@se.cuhk.edu.hk; hmmeng@se.cuhk.edu.hk). Corresponding author: Xunying Liu.

~~{Our code has been released on \href{https://github.com/AmourWaltz/BayesLMs}{https://github.com/AmourWaltz/BayesLMs}.}
}
}

\maketitle

\begin{abstract}
    State-of-the-art neural network language models (NNLMs) represented by long short term memory recurrent neural networks (LSTM-RNNs) and Transformers are becoming highly complex. They are prone to overfitting and poor generalization when given limited training data. 
    To this end, an overarching full Bayesian learning framework encompassing three methods is proposed in this paper to account for the underlying uncertainty in LSTM-RNN and Transformer LMs. 
    The uncertainty over their model parameters, choice of neural activations and hidden output representations are modeled using Bayesian, Gaussian Process and variational LSTM-RNN or Transformer LMs respectively. 
    {Efficient inference approaches were used to automatically select the optimal network internal components to be Bayesian learned using neural architecture search. 
    A minimal number of Monte Carlo parameter samples as low as one was also used.
    These allow the computational costs incurred in Bayesian NNLM training and evaluation to be minimized.}
    Experiments are conducted on two tasks: AMI meeting transcription and Oxford-BBC LipReading Sentences 2 (LRS2) overlapped speech recognition using state-of-the-art LF-MMI trained factored TDNN systems featuring data augmentation, speaker adaptation and audio-visual multi-channel beamforming for overlapped speech.
    Consistent performance improvements over the baseline LSTM-RNN and Transformer LMs with \revision{point estimated} model parameters and drop-out regularization were obtained across both tasks in terms of perplexity and word error rate (WER). 
    In particular, on the LRS2 data, statistically significant WER reductions up to {1.3\% and 1.2\% absolute (12.1\% and 11.3\% relative)} were obtained over the baseline LSTM-RNN and Transformer LMs respectively {after model combination \revision{between Bayesian NNLMs and their respective baselines}}.
\end{abstract}


%
\IEEEpeerreviewmaketitle

\section{Introduction}
\label{sec:Intro}
\IEEEPARstart{L}{anguage} models (LMs) are key components in many speech technology applications such as automatic speech recognition (ASR) systems. 
LMs are used to compute the probability of a word sequence {$\boldsymbol W =\left ( \boldsymbol w_0,\boldsymbol w_1,\dots,\boldsymbol w_n \right )$},
\begin{align}
\label{eq1}
    {{\small P(\boldsymbol W) = {P({{\boldsymbol w}_0,\boldsymbol w_1,...,{\boldsymbol w}_n})=\prod_{t=1}^{n}  P({{\boldsymbol w}_t|{\boldsymbol w}_{0},..., {\boldsymbol w}_{t-1}})}}}
\end{align}
which can be further expressed as the product of individual word probabilities conditioned on their respective preceding contexts\revision{, where $\boldsymbol w_0$ is typically a start symbol $<$s$>$, and generally $P(\boldsymbol w_0)=1$.}
The key research issue for statistical language modeling is to learn long-range contextual dependencies.
Directly modeling long-span word histories using conventional back-off $n$-gram models \cite{mgramlm} generally leads to a severe data sparsity issue \cite{chen1999empirical}.
To this end, over the past few decades significant efforts have been made to develop artificial neural network based language modeling techniques { \cite{bengio2003neural,schwenk2007continuous,arisoy2012deep,le2012structured,mikolov2010recurrent,sundermeyer2015feedforward,chen2016efficient,chen2019exploiting,irie2019language,li2020empirical,beck2020lvcsr,baquero2020improved,sun2021transformer,irie2020advancing,sheikh2021transformer}}.
Neural network language models (NNLMs) {that} represent longer span history contexts in a continuous and lower dimensional vector space can be used to improve generalization performance.

With the rapid progress of deep neural network (DNN) based ASR technologies in recent decades, the underlying model architectures of NNLMs have evolved from feedforward structures {\cite{bengio2003neural,schwenk2007continuous,arisoy2012deep,le2012structured}} to more advanced variants represented by long-short term memory recurrent neural networks (LSTM-RNNs) {\cite{mikolov2010recurrent,sundermeyer2015feedforward,chen2016efficient,chen2019exploiting}}, \cite{hochreiter1997long} and recently neural Transformers \cite{irie2019language,li2020empirical,beck2020lvcsr,baquero2020improved},\cite{vaswani2017attention} that are designed to model longer range contexts.
In particular, Transformer based NNLMs in recent years have defined state-of-the-art performance across a range of ASR tasks \cite{irie2019language,li2020empirical,beck2020lvcsr,baquero2020improved},\cite{xu2021mixed}. 
These models \cite{irie2019language,li2020empirical,beck2020lvcsr},\cite{xu2021mixed} are often constructed using a deep stacking of multiple self-attention based neural building blocks \cite{cheng2016long,lin2017structured,parikh2016decomposable}, 
each of which also includes residual connections \cite{he2016deep} and layer normalization modules \cite{ba2016layer}. 
Additional positional encoding layers \cite{vaswani2017attention,gehring2017convolutional} are used to augment the self-attention modules with word sequence order information. 
Performance improvements over conventional LSTM-RNN LMs have been reported \cite{irie2019language,zeyer2019comparison}. 

In state-of-the-art ASR systems based on both the conventional hybrid DNN-HMM architecture \cite{kingsbury2009lattice,vesely2013sequence,su2013error,povey2016purely,sak15sequence,michel2020frame} and the recent end-to-end (E2E) modeling paradigm represented by listen, attend and spell (LAS) \cite{chan2016listen}, connectionist temporal classification (CTC) \cite{graves2006connectionist}, RNN transducers (RNN-T) \cite{graves2012sequence} and Transformers \cite{guo2021recent}, the use of separate or externally fused LSTM-RNN or Transformer based language models { \cite{chen2016efficient,chen2019exploiting,irie2019language,li2020empirical,beck2020lvcsr,wang2020transformer,kim2021improved,meng2021internal,tuske2021limit}} that can benefit from the use of more text data in addition to the speech audio transcripts is essential.

However, the highly complex neural architecture designs of LSTM-RNN and Transformer LMs often lead to a large increase in the overall system complexity. 
When given limited training data of a target task domain, the use of \revision{point estimate based}, deterministic model parameters and neural structures in LSTM-RNN and Transformer LMs lead to a most salient challenge of handling model uncertainty, and mitigating the risk of over-fitting and poor generalization. 
One popular approach to address this issue in neural language models, and many other deep learning systems, is based on the dropout method \cite{baldi2013understanding,srivastava2014dropout,gal2016theoretically,gal2016dropout}, a simple and effective regularization approach.
Similar approaches based on, for example, weight noise \cite{murray1994enhanced,braun2019parameter,bishop1995training}, inject random noise directly into neural network parameters, \revision{or the incorporation of an additional L1 \cite{tibshirani1996regression} or L2 \cite{tikhonov1963solution} penalty term or maximum a posteriori (MAP) estimation \cite{chien2014bayesian,chien2015bayesian} into the training cost function} to improve generalization. 
However, {these intuitive regularization techniques do not provide a full Bayesian framework} to model the underlying uncertainty when estimating complex and over-parameterized DNN models on limited training data.

A more general solution considered in this paper to address the above model uncertainty issue for NNLMs is based on Bayesian neural network learning. 
In the machine learning community, Bayesian learning has been established as a well formulated framework to  account for the underlying uncertainty over model parameters \cite{neal2012bayesian,mackay1992practical,bishop2006pattern,graves2011practical,blundell2015weight}, 
or hidden layer output representations \cite{kingma2014auto,chung2015recurrent} in artificial neural network systems. 
Previous studies have further shown the commonly used dropout method can be formulated as special instance of Bayesian neural networks \cite{gal2016dropout}. 
Within the speech community, previous {research works} on Bayesian deep learning approaches were conducted mainly in the context of either acoustic model components of conventional hybrid DNN-HMM systems \cite{lam2018gaussian,hu2019lf,hu2019bayesian,hu2021bayesian}, or recently E2E architectures {\cite{braun2019parameter}}. 
They have also been successfully applied to speaker adaptation \cite{xie2019blhuc,xie2021bayesian} and speaker verification \cite{li2020bayesian} tasks. 

In contrast, limited previous {research works} conducted on Bayesian neural network language modeling approaches have been largely restricted to conventional RNN {\cite{chien2014bayesian,chien2015bayesian,gan2016scalable,fortunato2017bayesian}}, LSTM-RNN and gated recurrent unit (GRU) based NNLMs \cite{lam2019gaussian,yu2019comparative}. 
There are several issues associated with these prior {studies}. 
First, Bayesian learning approaches have not been studied in the context of Transformer LMs for speech technology applications. 
Instead, the only previous {works} in this direction \cite{tran2018bayesian,blt} were conducted on machine translation and probabilistic programming tasks. 
Second, these earlier studies focused on modeling the uncertainty over neural network parameters while lacking a holistic comparison against techniques designed to model the additional uncertainty associated with the underlying neural architectures and hidden layer output representations. 
Finally, when the Monte Carlo parameter sampling approach \cite{barber1998ensemble,kingma2014stochastic} commonly used for Bayesian neural network inference is applied to all individual layers of LSTM-RNN or Transformer LMs, 
the computational cost incurred in Bayesian model estimation grows exponentially with respect to the number of hidden layers in these systems. 
This leads to a major scalability issue that limits the practical application of Bayesian neural language modeling approaches. 

In order to address these issues, this paper proposes a mathematically well-defined full Bayesian learning framework to account for model uncertainty in both LSTM-RNN and Transformer LMs.
Three full Bayesian learning based approaches are proposed in this paper. 
The uncertainty over deterministic model parameters are modeled using Bayesian LSTM-RNN and Transformer LMs. 
The uncertainty over their underlying neural architecture design is further addressed using non-parametric Gaussian Process (GP) based neural activation functions. 
This leads to Gaussian Process LSTM-RNN and Transformer LMs. 
Variational LSTM-RNN and Transformer LMs are utilized to model the uncertainty over the hidden layer outputs in their respective conventional counterparts.
A complete and side by side comparison between these three Bayesian NNLM approaches is drawn. 
All the latent variable distributions considered in these three Bayesian methods are estimated using efficient variational inference based approaches. 
{Efficient inference approaches were used to automatically select the optimal network internal components to be estimated using a Bayesian approach and neural architecture search (NAS). 
A minimal number of Monte Carlo parameter samples as low as one was also used.}
{These allow the computational costs incurred in Bayesian NNLM training and evaluation to be minimized for both LSTM-RNN and Transformer LMs}
Experiments are conducted on two different tasks: a) AMI meeting transcription \cite{hain2006ami}; b) a multi-channel overlapped speech recognition task on the Oxford-BBC LipReading Sentences 2 (LRS2) corpus \cite{chung2017lip} using state-of-the-art LF-MMI trained factored TDNN systems \cite{povey2016purely} featuring speed perturbation, i-Vector \cite{dehak2010front,madikeri2016implementation,saon2013speaker} and learning hidden unit contribution (LHUC) based speaker adaptation \cite{swietojanski2014learning}, in addition to the use of audio-visual multi-channel beamforming for overlapped speech recognition. 
Consistent performance improvements over the baseline LSTM-RNN and Transformer LMs using \revision{point estimated} model parameters and hidden activation configurations as well as drop-out regularization were obtained across both tasks in terms of perplexity and word error rate (WER). 
In particular, on the LRS2 multi-channel overlapped speech recognition task, statistically significant average WER reductions of {1.3\% and 1.2\% absolute (12.1\% and 11.3\% relative)} 
were obtained over the baseline LSTM-RNN and Transformer LMs respectively after model combination \revision{between Bayesian NNLMs and their respective baselines}.
{A signal-to-noise ratio (SNR) computed over the variational Gaussian distribution mean introduced in \cite{braun2019parameter,blundell2015weight} was also used to measure the parameter uncertainty in the proposed Bayesian NNLMs.}

The main contributions of this paper are summarized below: 

1) To the best of our knowledge, this paper is the first work to systematically investigate a mathematically well grounded, full Bayesian framework to account for the underlying uncertainty over model parameters, neural activations and hidden layer output representations in both LSTM-RNN and Transformer based LMs. 
In contrast, prior {research works} on Bayesian neural network language modeling approaches were limited to conventional RNN \cite{chien2014bayesian,chien2015bayesian,gan2016scalable,fortunato2017bayesian}, LSTM-RNN and GRU based NNLMs \cite{lam2019gaussian,yu2019comparative}. 
In this paper, a complete and side by side contrast between these three Bayesian neural language modeling approaches across multiple speech recognition tasks is drawn and serves to provide insights on how to design practical Bayesian LSTM-RNN and Transformer LMs.  

2) The research presented in this paper also presents the first investigation of Gaussian Process and variational Transformer LMs published to date to account for additional uncertainty over neural activations and hidden layer output representations. 
In contrast, the prior {research work} on Bayesian Transformer LMs \cite{xue2021bayesian} considered modeling parametric uncertainty only.

3) Efficient inference algorithms are proposed for various Bayesian learning based LSTM-RNN and Transformer LMs. 
In addition to the use of a minimal number of Monte Carlo parameter samples drawn as low as one, a novel neural architecture search based method is proposed to automatically locate the most important network internal components to be Bayesian estimated, for example, a small number of lower positioned Transformer layers exhibiting larger uncertainty than others due to higher variability in their respective data inputs. 
This allows the model training costs for various Bayesian LSTM-RNN and Transformer LMs to {be comparable} to those of conventional NNLMs with \revision{point estimate based} parameters, and improves their scalability when deeper model architectures are used. 
In contrast, previous {research works} \cite{yu2019comparative,xue2021bayesian} manually selected a subset of hidden layers for Bayesian estimation, while an exhaustive search over all neural network component combinations was computationally infeasible.

The rest of this paper is organized as follows. 
Section \ref{sec:NNLMs} reviews the conventional LSTM-RNN and Transformer based LMs. 
Section \ref{sec:BLN} presents three full Bayesian learning approaches for LSTM-RNN and Transformer LMs. 
The accompanying set of implementation issues to improve their efficiency are presented in section \ref{sec:Imp}. 
Experiments and results are shown in section \ref{sec:exp}.
Finally, conclusions are drawn and future works discussed in section \ref{sec:con}.

\section{Neural Network Language Models}
\label{sec:NNLMs}
This section reviews neural network language models based on the LSTM-RNN and Transformer architectures.
\vspace{-0.6em}

\subsection{LSTM-RNN Language Models}
\label{sec:lstm}
\begin{figure}[!ht]
\newcommand{\tabincell}[2]{\begin{tabular}{@{}#1@{}}#2\end{tabular}}
\setlength{\parskip}{0.0ex}
  \centering
  \centerline{\includegraphics[width=8.8cm]{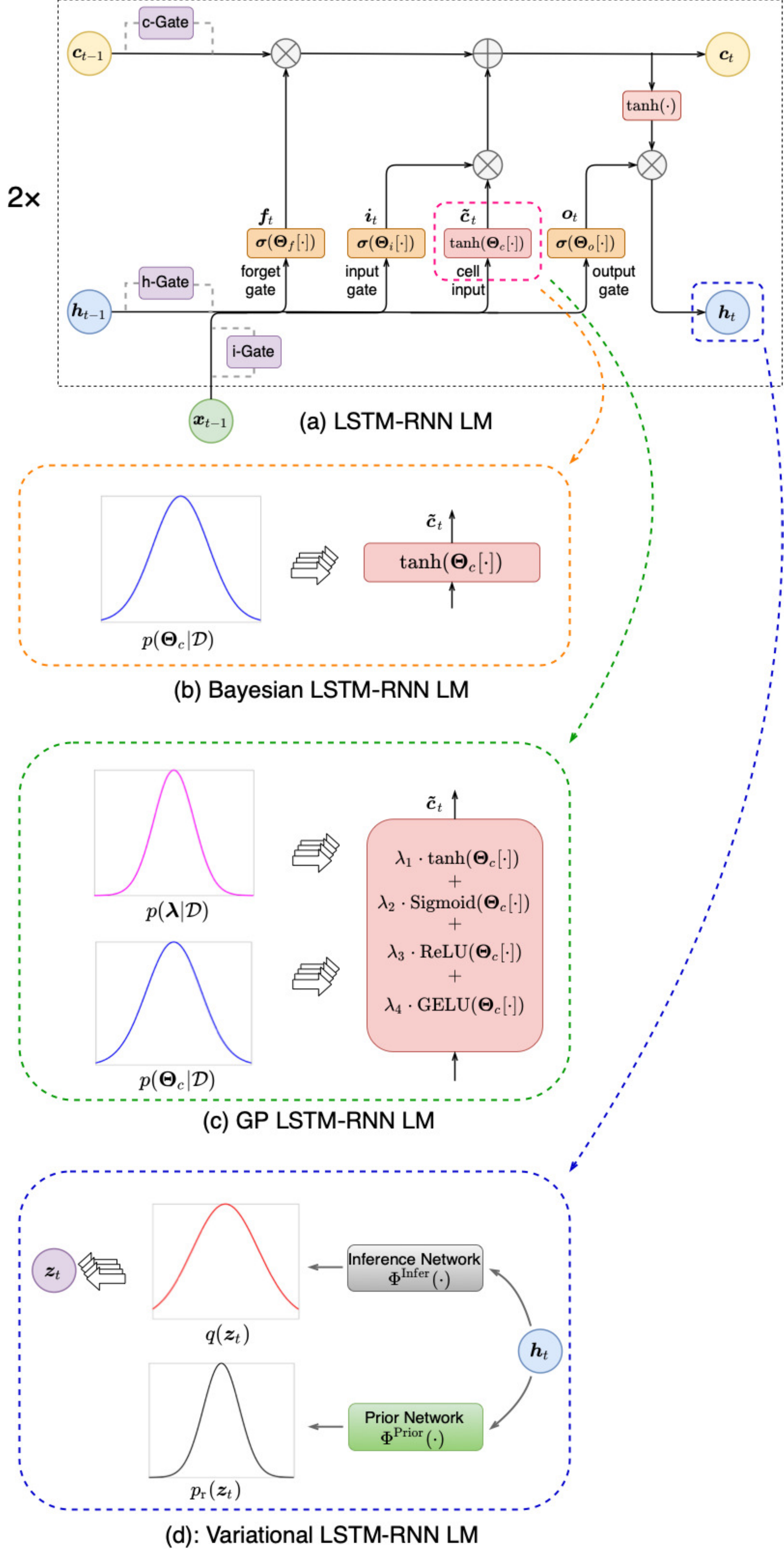}}
\captionsetup{font={small,it}}
\caption{Examples of 
(a) a standard LSTM-RNN LM with {point estimated} parameters and two hidden layers, where $\boldsymbol \Theta_i$, $\boldsymbol \Theta_f$, $\boldsymbol \Theta_c$, $\boldsymbol \Theta_o$ denote the respective weight parameters for {the input, forget, output gates and cell input} of an LSTM unit shown in Equation (\ref{eq4})-(\ref{eq7});
{(b) a Bayesian LSTM-RNN LM} with latent Gaussian posterior distributions modeling parametric uncertainty on the {cell input}; 
{(c) a Gaussian Process LSTM-RNN LM} with an interpolation over multiple basis activation functions, where latent Gaussian posterior distributions are used to model uncertainty over both the basis coefficients and activation function parameters on the {cell input};  
and (d) a variational LSTM-RNN LM using latent Gaussian posterior distributions to model the uncertainty over hidden layer outputs.}
\vspace{-1.5em}
\label{fig:lstm}
\end{figure}

In conventional LSTM-RNN language models, the word probability is computed as:
\vspace{-0.3em}
\begin{align}
\label{eq2}
    {\small P({\boldsymbol w}_t|{{\boldsymbol w}_{0:t-1}}) \approx P({\boldsymbol w}_t|{\boldsymbol h}_{t-1},{\boldsymbol w}_{t-1}) = P({\boldsymbol w}_t|{\boldsymbol h}_{t})}
\end{align}
where {${\boldsymbol w}_{0:t-1}$} denotes the word sequence {$\left (\boldsymbol w_{0}, \dots, \boldsymbol w_{t-1} \right )$} and $\boldsymbol h_{t-1}\in {\mathbb R}^{\it D}$ is the hidden state that encodes the previous word sequence {$\left ( \boldsymbol w_{0}, \dots, \boldsymbol w_{t-2} \right )$} into a continuous vector representation, and {\it D} is the number of hidden nodes. In NNLMs, the most recent history ${\boldsymbol w_{t-1}}$ is represented by a one-hot vector ${\boldsymbol{\tilde w}_{t-1}} \in {\mathbb R}^{\it N}$ as the input, where {\it N} is the vocabulary size. 
Generally, the LSTM-RNN LM contains three parts: the word embedding layer, the recurrent layer and the output layer. In order to address data sparsity, the word embedding layer {projects the one-hot word ${\boldsymbol{\tilde w}_t}$ as the input vector into a neural network. After being forward fed through non-linear transformations contained in multiple hidden layers, the one-hot word ${\boldsymbol{\tilde w}_t}$ is then projected into a continuous space} ${\boldsymbol{x}_t} \in {\mathbb R}^{\it M}$, where {\it M} is the embedding size and usually ${\it M} \ll {\it N}$:
\vspace{-0.3em}
\begin{align}
\label{eq3}
    {\small {{{{\boldsymbol x}_t={\boldsymbol \Theta}_{\mathrm U}\boldsymbol{\tilde w}_t}}}}
\end{align}
where ${\boldsymbol \Theta}_{\mathrm U} \in {\mathbb R}^{\it M \times N}$ is a projection matrix that can be learned during training process before being further fed into the hidden layers.
After word embedding, the hidden state $\boldsymbol h_t$ is obtained within the {\it memory cell} in the LSTM architecture \cite{hochreiter1997long} where the information of previous hidden state $\boldsymbol h_{t-1}$ is combined with the word $\boldsymbol w_{t-1}$.
As shown in Fig.\ref{fig:lstm} (a), at time step $t$ the respective outputs from the gates are obtained, including the forget gate ${\boldsymbol f_{t}}$, the input gate ${\boldsymbol i_{t}}$, the {cell input} ${\boldsymbol {\tilde c}_{t}}$ and the output gate ${\boldsymbol o_{t}}$. 
These gating functions are used to control the information flow within the cells to store the historical contexts over longer time steps and address the vanishing gradient issue {with the additive connection in Equation (\ref{eq8})} found in conventional RNN LMs. 
The respective outputs from these gates are calculated by
\vspace{-0.4em}
\begin{align}
    \label{eq4}
    {\small {\boldsymbol f}_{t}}&{\small ={\boldsymbol \sigma}\left ( {{{\boldsymbol \Theta}_{f} \left [ {{\boldsymbol x}_{t-1}}^\top, {{\boldsymbol h}_{t-1}}^\top, 1 \right ]^\top}} \right )}\\
    \label{eq5}
    {\small {\boldsymbol i}_{t}}&{\small ={\boldsymbol \sigma}\left ( {{{\boldsymbol \Theta}_{i} \left [ {{\boldsymbol x}_{t-1}}^\top, {{\boldsymbol h}_{t-1}}^\top, 1 \right ]^\top}} \right )}\\
    \label{eq6}
    {\small {\boldsymbol {\tilde c}}_{t}}&={\small {\mathbf {tanh}}\left ( {{{\boldsymbol \Theta}_{c} \left [ {{\boldsymbol x}_{t-1}}^\top, {{\boldsymbol h}_{t-1}}^\top, 1 \right ]^\top}} \right )}\\
    \label{eq7}
    {\small {\boldsymbol o}_{t}}&{\small ={\boldsymbol \sigma}\left ( {{{\boldsymbol \Theta}_{o} \left [ {{\boldsymbol x}_{t-1}}^\top, {{\boldsymbol h}_{t-1}}^\top, 1 \right ]^\top}} \right )}
\end{align}
where ${\boldsymbol \sigma}$ denotes the Sigmoid activation function and ${\boldsymbol \Theta}_{f},{\boldsymbol \Theta}_{i},{\boldsymbol \Theta}_{c},{\boldsymbol \Theta}_{o} \in {\mathbb R}^{\it {D \times (M + D + 1)}}$ denote the weight parameter matrices associated with the above gates respectively. 
The final hidden state is normally computed recursively from left to right in uni-directional LSTM-RNN LMs
{\cite{mikolov2010recurrent,sundermeyer2015feedforward}}. Given the gating outputs, the hidden state ${\boldsymbol h}_t$ and cell memory ${\boldsymbol {c}}_{t}$ are:
\vspace{-0.4em}
\begin{align}
    \label{eq8}
    &{\small {\boldsymbol {c}}_{t} = {\boldsymbol f_t} \circ {\boldsymbol {c}}_{t-1} + {\boldsymbol i}_{t} \circ {\boldsymbol {\tilde c}}_{t}}\\
    \label{eq9}
    &{\small {\boldsymbol h}_t = {\boldsymbol o}_t \circ {\mathbf {tanh}} \left ( {\boldsymbol {c}}_{t} \right )}
\end{align}
\revisions{where ${{\circ}}$ is the Hadamard product.}
Finally, the output layer uses the hidden state vector ${\boldsymbol h_{t}}$ to compute the word probabilities via a {\it Softmax} function:
\vspace{-0.5em}
\begin{align}
    \label{eq10}
    {\small P\left ( {\boldsymbol w}_t\mid {\boldsymbol h}_t \right ) = \frac{\exp {\left ( {{{{\boldsymbol \Theta}_{\rm v}^{\left ( {\boldsymbol w}_t\right )}}  {\boldsymbol h}_t}} \right )}}{\sum_{{\boldsymbol w}\in \mathcal V}\exp {\left ( {{{\boldsymbol \Theta}_{\rm v}^{\left ( {\boldsymbol w}\right )}  {\boldsymbol h}_t}} \right )}}}
\end{align}
where {${\boldsymbol \Theta}_{\rm v}^{\left ( {\boldsymbol w}\right )}\in \mathbb R^{1\times D}$} is the weight vector for word ${\boldsymbol w}$ in the output layer, and $\mathcal V$ represents the vocabulary.

\vspace{-0.4em}

\begin{figure}[!ht]
\newcommand{\tabincell}[2]{\begin{tabular}{@{}#1@{}}#2\end{tabular}}
\setlength{\parskip}{0.0ex}
  \centering
  \centerline{\includegraphics[width=8.8cm]{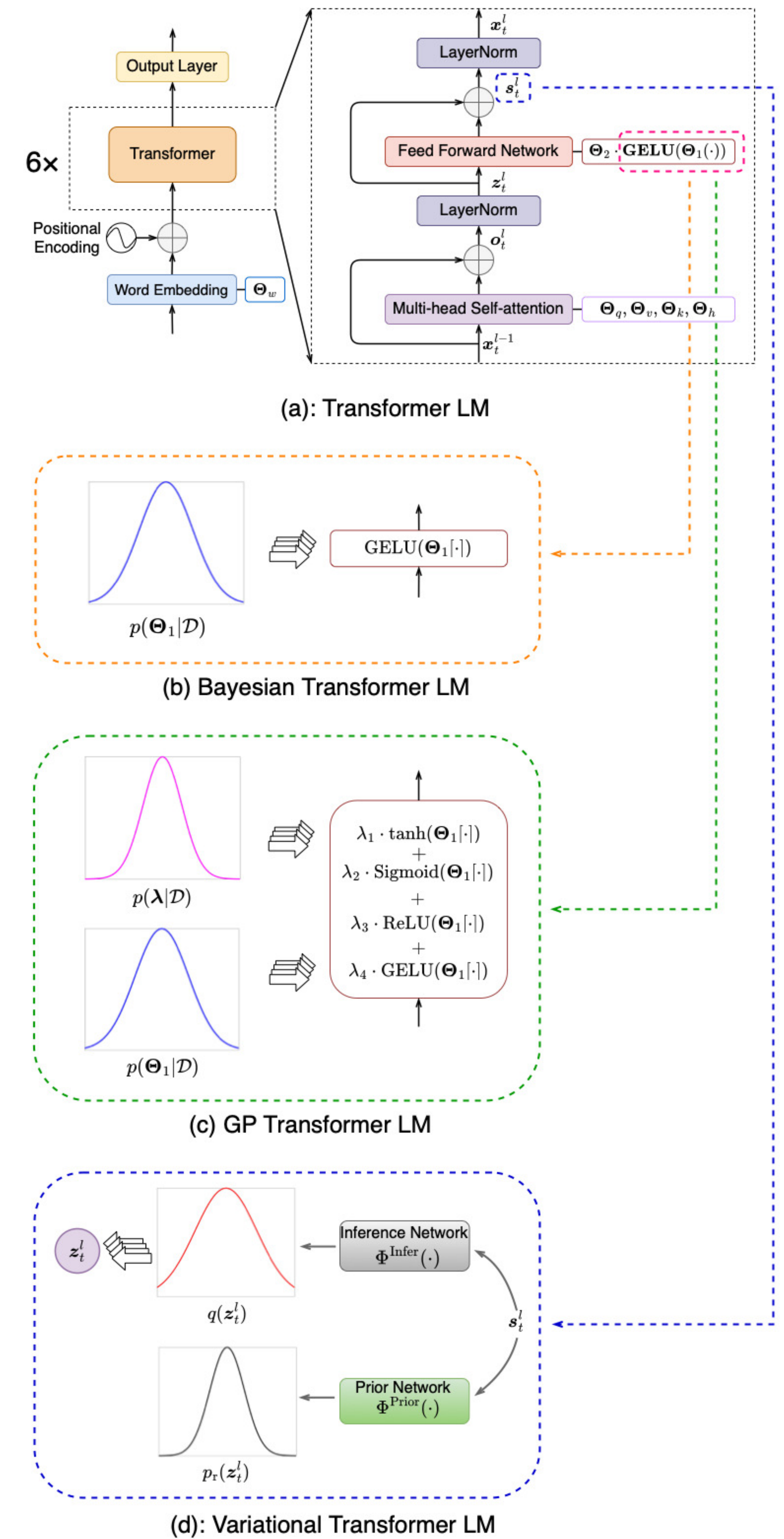}}
\captionsetup{font={small,it}}
\caption{Examples of 
(a) a standard Transformer LM with {point estimated} parameters and six layers, where $\boldsymbol \Theta_w$, $\boldsymbol \Theta_q$, $\boldsymbol \Theta_k$, $\boldsymbol \Theta_v$, $\boldsymbol \Theta_h$, $\boldsymbol \Theta_1$, $\boldsymbol \Theta_2$ denote the parameters of the word embedding layer, multi-head self-attention and feed forward network modules in Equation (\ref{eq11}), (\ref{eq13}) and (\ref{eq15});
{(b) a Bayesian Transformer LM} with latent Gaussian posterior distributions modeling parametric uncertainty for a feed-forward sub-network; 
{(c) a Gaussian Process Transformer LM} with an interpolation over multiple basis activation functions, where latent Gaussian posterior distributions are used to model uncertainty over both the basis coefficients and activation function parameters for a feed-forward sub-network;  
and (d) a variational Transformer LM using latent Gaussian posterior distributions to model the uncertainty over hidden layer outputs.}
\vspace{-1.7em}
\label{fig:trans}
\end{figure}

\subsection{Transformer Language Models}
\label{sec:trans}
The Transformer model architecture considered in this paper features a deep stacking of multiple Transformer decoder blocks.
As shown in Fig.\ref{fig:trans} (a), each Transformer decoder block consists of a multi-head self-attention \cite{cheng2016long,lin2017structured,parikh2016decomposable} and feed forward neural network modules in each block. 
Residual connections \cite{he2016deep} and layer normalization \cite{ba2016layer} are also inserted between the two modules. 
Assume that ${\boldsymbol x}_t^{l-1}\in {\mathbb R}^M$ represents the output of the $(l-1)$-th Transformer layer at time step $t$, where $M$ is the dimensionality of word embedding. In the $l$-th layer, the multi-head self-attention module transforms ${\boldsymbol x}_t^{l-1}$ to ${\boldsymbol z}_t^{l}$ as follows:
\begin{align}
    \label{eq11}
    {\small {\boldsymbol q}_{t}^{l}, {\boldsymbol k}_{t}^{l},} &{\small {\boldsymbol v}_{t}^{l}={{\boldsymbol \Theta}^l_q\left [ {{\boldsymbol x}_{t}^{l-1}}^{\top},1 \right ]^\top,{\boldsymbol \Theta}^l_k\left [ {{\boldsymbol x}_{t}^{l-1}}^{\top},1 \right ]^\top,{\boldsymbol \Theta}^l_v\left [ {{\boldsymbol x}_{t}^{l-1}}^{\top},1 \right ]^\top}} \\
    &{\small {\boldsymbol y}_{t}^{l} = \text{\bf {{Attn}}}\left ({\boldsymbol k}_{1}^{l}, \dots, {\boldsymbol k}_{t}^{l}, {\boldsymbol v}_{1}^{l}, \dots, {\boldsymbol v}_{t}^{l}, {\boldsymbol q}_{t}^{l}\right )} \nonumber \\ 
    \label{eq12}
    &{\small = \text{\bf Softmax}\left ( \frac{{{{{\boldsymbol q}_{t}^{l}}^{\top}({\boldsymbol k}_{1}^{l}, \dots, {\boldsymbol k}_{t}^{l})}}}{ {{\sqrt{M}}} } \right )\left ({\boldsymbol v}_{1}^{l}, \dots, {\boldsymbol v}_{t}^{l} \right )} \\
    \label{eq13}
    &{\small \boldsymbol o^l_t = {{{\boldsymbol \Theta}_h^{l} \left [ {{\boldsymbol y}_{t}^{l}}^{\top},1 \right ]^\top}} + {\boldsymbol x}_{t}^{l-1}} \\
    \label{eq14}
    &{\small {\boldsymbol z}_{t}^{l} = \text{\bf LayerNorm}({\boldsymbol o}_{t}^{l})}
\end{align}
where ${\boldsymbol \Theta}^l_q,{\boldsymbol \Theta}^l_k,{\boldsymbol \Theta}^l_v\in {\mathbb R}^{M\times (M+1)}$ are projection matrices of $l$-th self-attention module that maps the input ${\boldsymbol x}_t^{l-1}$ into query ${\boldsymbol q}_t^l$, key ${\boldsymbol k}_t^l$ and value ${\boldsymbol v}_t^l$ respectively.
$\text{\bf {Attn}}(\cdot)$ represents the scaled multi-head dot product self-attention mechanism \cite{vaswani2017attention}.
{Only self-attention mechanism with a single head is used in Equation ({\ref{eq12}}).}
${\boldsymbol y}_t^l$ is the sequence of cached key-value vector pairs up to time step $t$, which can be restricted to only contain the history context information and thus prevent the model from using any future context, and ${\frac{1}{\sqrt{M}}}$ denotes the scaling factor.
$\text{\bf LayerNorm}(\cdot)$ is the layer normalization operation \cite{ba2016layer}.
${\boldsymbol \Theta}_h^{l}\in {\mathbb R}^{M\times (M+1)}$ is the projection matrix applied to the outputs of the $\text{{\bf Attn}}(\cdot)$ operation for residual connection \cite{he2016deep}. The normalized output ${\boldsymbol z}_{t}^{l}$ is then fed into the feed forward network module:
\vspace{-0.4em}
\begin{align}
    \label{eq15}
    {\small {\boldsymbol s}_{t}^{l}} &{\small ={{\boldsymbol \Theta}_2^{l}\left [ {\text{\bf GELU}\left ({\boldsymbol \Theta}_1^{l}\left [ {{\boldsymbol z}_{t}^{l}}^{\top},1 \right ]^\top\right )}^{\top},1\right ]^\top} + {\boldsymbol z}_{t}^{l}} \\
    \label{eq16}
    {\small {\boldsymbol x}_{t}^{l}} &= {\small \text{\bf LayerNorm}({\boldsymbol s}_{t}^{l})}
\end{align}
where ${\boldsymbol \Theta}_1^{l}\in {\mathbb R}^{D\times (M+1)},{\boldsymbol \Theta}_2^{l}\in {\mathbb R}^{M\times (D+1)}$ denote the weight matrices of the feed forward networks and $D$ is the size of hidden nodes. The Gaussian error linear unit $\text{\bf GELU}(\cdot)$ {\cite{dan2016bridging,radford2018improving}} is adopted as the activation function in this work.
\revision{The output of the last Transformer layer, $\boldsymbol x_t^l$, is fed into the output layer, where the word probabilities are computed as in Equation (10), assuming $\boldsymbol h_t=\boldsymbol x_t^l$.}

\vspace{-0.2em}


\section{Bayesian Learning based NNLMs}
\label{sec:BLN}
This section presents three Bayesian neural network language modeling approaches based on Bayesian neural networks, Gaussian process and variational neural networks. These are proposed in this paper to model the uncertainty over model parameters, hidden activation functions and their outputs respectively in LSTM-RNN and Transformer based NNLMs.

\vspace{-0.6em}

\subsection{Bayesian NNLMs}
\label{sec:BNNLM}
Conventional NNLMs using \revision{point estimate} parameters fail to account for the model uncertainty associated with the word prediction.
When given limited training data, {these NNLMs} are prone to over-ﬁtting and poor generalization.
To this end, Bayesian neural networks (BNNs) \cite{neal2012bayesian,mackay1992practical,bishop2006pattern,graves2011practical,blundell2015weight} can be used by modeling the uncertainty over model parameters ${\boldsymbol \Theta}$ using as a probability distribution $p_{\mathrm r}({\boldsymbol \Theta})$.
The model parameters $\boldsymbol \Theta$ include all the activation function parameters, for example, those of various LSTM unit gates $\boldsymbol \Theta_i$, $\boldsymbol \Theta_f$, $\boldsymbol \Theta_c$, $\boldsymbol \Theta_o$ of Equation (\ref{eq4})-(\ref{eq7}) in LSTM-RNN LMs, and the feed-forward layer parameters $\boldsymbol \Theta_1$ of Equation (\ref{eq15}) in Transformer LMs. The {log probability} of a given $n$ word sentence {$\boldsymbol W =\left ( \boldsymbol w_0,\boldsymbol w_1,\dots,\boldsymbol w_n \right )$} is obtained as:
\vspace{-0.5em}
\begin{align}
\label{eq17}
    {\small \log P}&{\small ({{{\boldsymbol W|\mathcal D}}})
    =\log \int P({\boldsymbol W}|{\boldsymbol \Theta})p({\boldsymbol \Theta} | {\cal D}) \mathrm d{\boldsymbol \Theta}} \nonumber\\
    &{\small =\log \int \prod_{t=1}^n P({\boldsymbol w_t}|{{\boldsymbol w_0,\dots\boldsymbol w_{t-1}},\boldsymbol \Theta})p({\boldsymbol \Theta} | {\cal D}) \mathrm d{\boldsymbol \Theta}}
\end{align}
\vspace{-0.7em}

\noindent where $p({\boldsymbol \Theta} | {\cal D})$ is the posterior distribution over parameters to be learned from the given $N$ word training data set {${\cal D} = ( {\boldsymbol w}_{0}, {\boldsymbol w}_{1}, ..., {\boldsymbol w}_{N})$}. 

\revisions{In classical Bayesian neural networks, including Bayesian NNLMs considered in this paper, the estimation of the latent variable distribution $p(\boldsymbol \Theta|\mathcal D)$ must be learned in order to perform inference on any unseen test data sentence $\boldsymbol W$ using the predictive distribution $P(\boldsymbol W |\mathcal D)$, for example, given in Equation (\ref{eq17}) for Bayesian NNLMs.
The latent variable distribution $p(\boldsymbol \Theta|\mathcal D)$ needs to be learned through maximizing the model evidence expressed as the training data marginal likelihood ($\!\!$\cite{bishop2006pattern}, Chapter 5.7). 
Since directly computing the evidence integral is intractable for neural networks, a range of approximated inference schemes were proposed in previous works including Laplace approximation \cite{bishop2006pattern,mackay1992practical}, Markov chain Monte Carlo \cite{neal2012bayesian,neal1994bayesian}, and the more recently proposed variational inference \cite{graves2011practical,kingma2014auto,barber1998ensemble,kingma2014stochastic} that is also used in this paper.}
The following variational lower bound of the marginal log-likelihood is maximized instead \cite{kingma2014auto}: 
\vspace{-0.4em}
\begin{align}
    \label{eq18}
    &{\small \log P({\cal D}) = \log\int P({\cal D}|{\boldsymbol \Theta})p_{{\mathrm r}}({\boldsymbol \Theta}) \mathrm d {\boldsymbol \Theta}} \\
    \label{eq19}
    \vspace{-0.2em}
    &{\small \ge \underbrace{\int \log P({\cal D}|{\boldsymbol \Theta})q({\boldsymbol \Theta}) \mathrm d {\boldsymbol \Theta}}_{\mathcal{L}_{1}} - \underbrace{\text{KL}\left [ q({\boldsymbol \Theta})||p_{\mathrm r}({\boldsymbol \Theta})\right ]}_{\mathcal{L}_{2}}}
\end{align}
\vspace{-0.6em}

\noindent where $q({\boldsymbol \Theta})$ is the variational approximation of the parameter posterior distribution $p({\boldsymbol \Theta}|\cal D)$. 
$p_{\mathrm r}({\boldsymbol \Theta})$ is the prior distribution of ${\boldsymbol \Theta}$ and $\mathrm{KL}\left [q({\boldsymbol \Theta})||p_{\mathrm r}({\boldsymbol \Theta})\right ]$ represents the Kullback-Leiber (KL) divergence between $q({\boldsymbol \Theta})$ and $p_{\mathrm r}({\boldsymbol \Theta})$. 
The exact computation of this KL regularization term is non-trivial for general forms of parameter posterior and prior distributions. 
For efficiency, \revisions{both $q({\boldsymbol \Theta})$ and $p_{\mathrm r}({\boldsymbol \Theta})$ are assumed to be diagonal Gaussian distributions \cite{graves2011practical,tran2018bayesian,blt,barber1998ensemble,kingma2014stochastic} in this work:}
\vspace{-0.4em}
\begin{align}
\label{eq20}
    \revisions{{\small {q({\boldsymbol \Theta})=\mathcal{N}({\boldsymbol \Theta}; {\boldsymbol\mu}, {\boldsymbol \Sigma}_{\sf diag}),\ \ \ \ p_{\mathrm r}({\boldsymbol \Theta})=\mathcal{N}({\boldsymbol \Theta}; {\boldsymbol \mu}^{\mathrm r}, {\boldsymbol \Sigma}^{\mathrm r}_{\sf diag})}}}
\end{align}
\vspace{-1.5em}

\noindent \revisions{where ${\boldsymbol \mu}$, ${\boldsymbol \Sigma}_{\sf diag}$ are the mean and diagonal covariance matrix for the variational distribution, while ${\boldsymbol \mu}^{\mathrm r}$, ${\boldsymbol \Sigma}^{\mathrm r}_{\sf diag}$ are those of the prior distribution.}
The first term of the marginal log-likelihood lower bound $\mathcal L_1$ is the expectation of log-likelihood of the word sequence ${\boldsymbol W}$ over the approximated posterior distribution $q({\boldsymbol \Theta})$.
This can be further efficiently approximated by the Monte Carlo sampling method: 
\vspace{-0.4em}
\begin{align}
\label{eq21}
   {\small \mathcal{L}_{1}} &{\small \approx\frac{1}{K}\sum_{k=1}^{K}\log P({\cal D}|{\boldsymbol \Theta}^{(k)})}\nonumber\\
   &{\small =\frac{1}{K}\sum_{k=1}^{K}\sum_{t=1}^{N}\log P({\boldsymbol w_t|{\boldsymbol w_0,\dots,\boldsymbol w_{t-1}}},{\boldsymbol \Theta}^{(k)})}
\end{align}

\vspace{-0.5em}

\noindent where $K$ is the \revision{number of samples} and ${\boldsymbol \Theta}^{(k)}$ is the $k$-th sample drawn from the distribution $q({\boldsymbol \Theta})$. 
It has been found that directly sampling ${\boldsymbol \Theta}^{(k)}$ using the {variational} distribution mean ${\boldsymbol \mu}$ and covariance matrix $\boldsymbol \Sigma_{\sf diag}$ is prone to instability during inference. 
To address this issue, the following re-parameterization method \cite{kingma2015variational} is adopted to sample \revisions{the $i$-th element ${\Theta_i}^{(k)}$ in the $k$-th sampled parameter ${\boldsymbol \Theta}^{(k)}$} as follows:
\vspace{-1.5em}
\begin{align}
\label{eq22}
    \revisions{\small {\Theta_i}^{(k)} = {\mu_i} + {{\sigma_i} \cdot {\epsilon_i}^{(k)}}, \ \ \ \ {\epsilon_i}^{(k)} \sim \mathcal{N(\mathrm 0, \mathrm{1})}}
\end{align}
\vspace{-1.6em}

\noindent \revisions{where $\mu_i$ is the {$i$-th} component of the $M$-dimensional mean vector $\boldsymbol\mu$, $\sigma_i$ is the square root of $\Sigma_{{\sf diag}, ii}$, the $i$-th diagonal element in the diagonal covariance matrix $\boldsymbol \Sigma_{\sf diag}$ of the variational distribution, i.e. ${\sigma_i}^{2} = \Sigma_{{\sf diag}, ii}$, and ${\epsilon_i}^{(k)}$ is the $k$-th sample in the total $K$ samples drawn for the $i$-th dimension.}
Assuming the independence between the model parameters of different NNLM internal components, for example, those of different gates of an LSTM cell unit, the above sampling needs to independently performed for the latent variational distribution associated with their respective parameters.

The second part $\mathcal L_2$ of Equation (\ref{eq19}) is the KL divergence between $q({\boldsymbol \Theta})$ and $p_{\mathrm r}({\boldsymbol \Theta})$. 
Under the Gaussian assumption, an analytical {closed-form} can be derived as: 
\vspace{-0.4em}
\begin{align}
\label{eq23}
    {\small \text{KL}\left [q({\boldsymbol \Theta})||p_{\mathrm r}({\boldsymbol \Theta})\right ] = 
    {{\sum_{i=1}^{M}}}\left \{ \log \frac{\sigma^{\mathrm r}_i}{\sigma_i}+\frac{{\sigma_i}^2+({\mu}_i-\mu^{\mathrm r}_i)^2}{2{\sigma^{\mathrm r}_i}^2} -\frac{1}{2}\right \} }
\end{align}
\vspace{-1.2em}

\noindent \revisions{where $\mu^{\mathrm r}_i$ and $\sigma^{\mathrm r}_i$ are the $i$-th component of the mean $\boldsymbol\mu^{\mathrm r}$ and the square root of $\Sigma^{\mathrm r}_{{\sf diag}, ii}$, the $i$th diagonal variance element of the prior distribution, i.e. ${\sigma^{\mathrm r}_i}^{2} = \Sigma^{\mathrm r}_{{\sf diag}, ii}$.}

The combined use of both the parameter sampling of Equation (\ref{eq21})-(\ref{eq22}), and the {closed-form} KL regularization term of Equation (\ref{eq23}) under the Gaussian assumption over the parameter posterior and prior distributions, 
allows the variational lower bound of Equation (\ref{eq19}) to be differentiable with respect to the variational distribution hyper-parameters $\boldsymbol\mu$ and \revisions{$\boldsymbol\Sigma_{\sf diag}$}, 
and can be fully integrated into the conventional back-propagation algorithm during Bayesian LSTM-RNN or Transformer LM training. 
\vspace{-0.5em}

\subsection{Gaussian Process based NNLMs}
\label{sec:GPLM}
Gaussian Processes (GPs) \cite{rasmussen2003gaussian} are non-parametric distributions over continuous functions in probabilistic modeling for many machine learning applications including regression and classification tasks and beyond. 
A function {$f(\cdot)$} modeled using Gaussian Process is represented as:
\vspace{-0.5em}
\begin{align}
    \label{eq24}
    &{\small f(\boldsymbol x)\sim \mathcal {GP} \left ( m(\boldsymbol x), k(\boldsymbol x,\boldsymbol x^\prime) \right )}
\end{align}
\vspace{-1.4em}

\noindent
where $\boldsymbol x,\boldsymbol x^\prime \in \mathbb R^M$ are an arbitrary pair of input data vectors, {and $f(\boldsymbol x) \in \mathbb R^D$ is a Gaussian distribution specified by the mean function $m(\boldsymbol x)$ and the kernel function $k(\boldsymbol x, \boldsymbol x^\prime)$ 
}. 
The above formulation is known as the kernel space view of GP models \cite{rasmussen2003gaussian}. 
The associated computational complexity over the kernel covariance function during inference is determined by the size of the training data set,
thus impractical when applied to large scale tasks, for example, language models trained on millions, or billions of words. 
An alternative and computationally more tractable form of GP uses basis function interpolation (\cite{rasmussen2003gaussian}, Chapter 2), also known as the following weight space view of GP, 
\vspace{-0.9em}
\begin{align}
    \label{eq25}
    {\small f(\boldsymbol x)=\boldsymbol \lambda^\top\boldsymbol\cdot {\boldsymbol \phi}(\boldsymbol x)={{\sum_{j=1}^{K}}}\lambda^j\phi^j(\boldsymbol x)}
\end{align}
\vspace{-1.2em}

\noindent where $k(\cdot,\cdot)= {\boldsymbol \phi}(\cdot)^{\top} {\boldsymbol \phi}(\cdot)$, $\boldsymbol \lambda$ denotes {the vector including $K$ basis coefficients} of different basis functions $\phi^j(\boldsymbol x)$ in ${\boldsymbol \phi}(\cdot)$.

A series of prior {research works} were conducted in the machine learning community to build the connection between neural networks and Gaussian processes. 
Neal \cite{1996Priors} proved that Bayesian neural networks \cite{mackay1992practical} with a single hidden layer of infinite width are equivalent to Gaussian Processes. 
Hazan and Jaakkola \cite{hazan2015steps} and later Lee \cite{lee2017deep} proposed the use of GP kernels to approximate infinitely wide deep neural networks. 
Such connection was further studied in deep Gaussian process (DGP) \cite{damianou13a} models where deep belief neural network layers were replaced by Gaussian Processes. 

The form of the traditional Bayesian neural networks introduced in Sec.\ref{sec:BNNLM} only considers the uncertainty associated with model parameters, but not the network structural configurations. 
For example, the choice over the activation functions that are widely used in Equation (\ref{eq4})-(\ref{eq7}), (\ref{eq9}) and (\ref{eq15}), can be learned using a learnable weighted interpolation in Equation (\ref{eq25}) over commonly used basis activation functions, i.e Sigmoid, tanh, ReLU and GELU, as considered in this paper. 
This interpolated form of hidden activation function, for example, the activation function in Equation (\ref{eq4}), is given by
\vspace{-0.7em}
\begin{align}
    \label{eq26}
    {\small \boldsymbol f_t = {{\sum_{j=1}^{K}}}\lambda^j \phi^j\left ( {{{\boldsymbol \Theta}_{f} \left [ {{\boldsymbol x}_{t-1}}^\top, {{\boldsymbol h}_{t-1}}^\top, 1 \right ]^\top}} \right )}
\end{align}
\vspace{-1.2em}

\noindent where ${\boldsymbol \Theta}_{f}$ is the activation function parameters and $\lambda^j$ is the $j$-th basis activation coefficient.

In addition to treating model parameters inside the activation functions as random variables as Bayesian NNLMs, 
further considering the uncertainty \revision{over} the basis activation coefficients $\boldsymbol\lambda$ leads to a more general form of Gaussian Process neural networks that simultaneously account for the uncertainty over both the NNLM model parameters and the choice of hidden activation functions. 
In GP NNLMs, assuming that the activation functions parameters $\boldsymbol \Theta$ {(linear weight parameters that are applied to the inputs of a neuron, before any non-linearity)} and basis activation coefficients $\boldsymbol\lambda$ are statistically independent of each other, the sentence level probability of a given word sequence $\boldsymbol W$ is then computed as follows: 
\vspace{-0.3em}
\begin{align}
    &{\small \log P({{\boldsymbol W|\mathcal D}}) =\log \int \!\!\!\int P({\boldsymbol W}|{\boldsymbol \Theta},\boldsymbol\lambda)p({\boldsymbol \Theta} | {\cal D})p({\boldsymbol \lambda} | {\cal D}) \mathrm d{\boldsymbol \Theta} \mathrm d{\boldsymbol \lambda}} \nonumber\\
    \label{eq27}
    &{\small = \log \int \!\!\! \int \prod_{t=1}^n P(\boldsymbol w_t|{\boldsymbol w_0,\dots,\boldsymbol w_{t-1}},\boldsymbol \Theta,\boldsymbol\lambda) p({\boldsymbol \Theta} | {\cal D})p({\boldsymbol \lambda} | {\cal D}) \mathrm d{\boldsymbol \Theta} \mathrm d{\boldsymbol \lambda}}
\end{align}
where $p({\boldsymbol \Theta} | {\cal D})$ and $p({\boldsymbol \lambda} | {\cal D}) $
denote the posterior probability distributions over the basis activation function parameters $\boldsymbol \Theta$ and the basis coefficients $\boldsymbol \lambda$ respectively, both to be learned from the training data ${\cal D}$.

Similar to the variational inference approach considered in Sec.\ref{sec:BNNLM} for Bayesian NNLMs, the following variational lower bound of marginal log-likelihood is maximized: 
\vspace{-0.2em}
\begin{align}
    {\small \log P({\cal D})} &{\small \ge \underbrace{\int\int \log P({\cal D}|{\boldsymbol \Theta},{\boldsymbol \lambda})q({\boldsymbol \Theta})q({\boldsymbol \lambda}) \mathrm d {\boldsymbol \Theta}\mathrm d {\boldsymbol \lambda}}_{\mathcal{L}_{1}}}\nonumber \\
    \label{eq28}
    &{\small - \underbrace{\text{KL}\left [q({\boldsymbol \Theta})||p_{\mathrm r}({\boldsymbol \Theta})\right ]}_{\mathcal{L}_{2}}- \underbrace{\text{KL}\left [q({\boldsymbol \lambda})||p_{\mathrm r}({\boldsymbol \lambda})\right ]}_{\mathcal{L}_{3}}}
\end{align}
where $q({\boldsymbol \Theta})$ and $q({\boldsymbol \lambda})$ are variational approximations of posterior distributions $p({\boldsymbol \Theta}|\cal D)$ and $p({\boldsymbol \lambda}| \cal D)$ that separately model the uncertainty over model parameters and the choice of hidden neural activation functions. 
Their respective KL-divergence against their prior distributions in ${\mathcal L_2}$ and ${\mathcal L_3}$ serve as regularization terms during GP NNLM inference.

In common with Bayesian NNLMs, both the prior distributions, $p_{\mathrm r}({\boldsymbol \Theta})$ and $p_{\mathrm r}({\boldsymbol \lambda})$, and the variational distributions, $q({\boldsymbol \Theta})$ and $q({\boldsymbol \lambda})$, for model parameters and basis activation coefficients are assumed to be Gaussian distributions \cite{tran2018bayesian,blt,barber1998ensemble,kingma2014stochastic} to produce {closed-form}, analytical solutions to the two KL terms.

The expectation of log-likelihood in the first term $\mathcal L_1$ of the {variational} lower bound in Equation (\ref{eq28}) is calculated using sampling method that requires activation parameter samples and basis activation coefficient samples to be drawn from their respective {variational} distributions, $q({\boldsymbol \Theta})$ and $q({\boldsymbol \lambda})$, with the re-parameterization method of Equation (\ref{eq22}).
\vspace{-0.3em}


\subsection{Variational NNLMs}
\label{sec:VLM}
In contrast to the BNNs and GPNNs presented in Sec.\ref{sec:BNNLM} and \ref{sec:GPLM}, variational neural networks (VNNs) \cite{kingma2014auto,chung2015recurrent,kuo2017variational,yu2019comparative} can be used to model uncertainty associated with hidden representations. 
Instead of modeling the uncertainty over the weight parameters $\boldsymbol \Theta$ inside the activation functions in BNNs or assuming additional uncertainty over the activation basis coefficients $\boldsymbol \lambda$ in GPNNs, VNNs introduce a latent variable $\boldsymbol z_t$ to encode the uncertainty associated with, for example, the hidden node output vector $\boldsymbol h_{t}$ of LSTM-RNNs in Equation (\ref{eq9}), or the outputs of feed-forward networks $\boldsymbol s_t^l$ of Transformer models in Equation (\ref{eq15}). 
The general form of sentence level probability of a given word sequence {$\boldsymbol W=\left ( \boldsymbol w_0,\boldsymbol w_1,\dots,\boldsymbol w_n \right )$} under variational NNLMs is computed as:
\vspace{-0.4em}
\begin{align}
    \label{eq29}
    {\small \log P}&{\small ({{{\boldsymbol W|\mathcal D}}})=\log \prod_{t=1}^n P({{\boldsymbol w_t|\boldsymbol w_0,\dots,\boldsymbol w_{t-1},\mathcal D}}}) \\
    &{\small {{\approx} }\log\prod_{t=1}^{n} \int P({\boldsymbol w_t}|{\boldsymbol w_0,\dots,\boldsymbol w_{t-1}},\boldsymbol z_t)p(\boldsymbol z_t | \boldsymbol h_t, {\cal D}) \mathrm d{\boldsymbol z_t}} \nonumber
\end{align}
\vspace{-0.6em}

\noindent where $p(\boldsymbol z_t | \boldsymbol h_t, {\cal D})$
denotes the latent variable distribution over neural network hidden layer outputs $\boldsymbol z_t$ to be learned from the training data {${\cal D} = \left ( \boldsymbol w_0, \boldsymbol w_1, ..., \boldsymbol w_{N} \right )$}, and the hidden layer vector outputs over time are {$\left ( \boldsymbol z_0,\boldsymbol z_1,\dots,\boldsymbol z_t \right )$}. 

During {variational} NNLMs training, the following training data marginal log-likelihood lower bound is maximized,
\vspace{-0.3em}
\begin{align}
    {\small \log P({\cal D})} &{\small \ge \sum_{t=1}^{N} \underbrace{\int \log P({\boldsymbol w}_t|{\boldsymbol w_{0},\dots,\boldsymbol w_{t-1}},\boldsymbol z_{t})q({\boldsymbol z_t}) \mathrm d {\boldsymbol z_t}}_{\mathcal{L}_{1}}}\nonumber\\
    \label{eq30}
    &{\small - \underbrace{\text{KL}\left [q({\boldsymbol z_t})||p_{\mathrm r}({\boldsymbol z_t})\right ]}_{\mathcal{L}_{2}}}
\end{align}
\vspace{-0.9em}

\noindent where $q({\boldsymbol z_t})$ is the variational approximation of $p(\boldsymbol z_t | \boldsymbol h_t, {\cal D})$, the posterior distribution over the hidden output vector $\boldsymbol z_t$, and $p_{{\mathrm r}}({\boldsymbol z_t})$ is its prior distribution. 
In common with both Bayesian and GP NNLMs, for efficiency during inference, both $q({\boldsymbol z_t})$ and $p_{{\mathrm r}}({\boldsymbol z_t})$ are assumed to be Gaussian distributions:
\vspace{-0.5em}
\begin{align}
    \label{eq31}
    \revisions{\small q({\boldsymbol z_t})=\mathcal{N}({\boldsymbol z_t}; {\boldsymbol\mu_t}, {\boldsymbol \Sigma}_{{\sf diag}, t}),\ \ \ \ p_{\mathrm r}({\boldsymbol z_t})=\mathcal{N}({\boldsymbol z_t}; {\boldsymbol \mu^{\mathrm r}_{t}}, {\boldsymbol \Sigma}^{\mathrm r}_{{\sf diag}, t})}
\end{align}
\vspace{-1.4em}

\noindent to allow the KL divergence between $q({\boldsymbol z_t})$ and $p_{{\mathrm r}}({\boldsymbol z_t})$ in Equation (\ref{eq30}) to be computed in a tractable form.
\vspace{-0.3em}
\begin{align}
    {\small \text{KL}}&{\small \left [q({\boldsymbol z_t})||p_{\mathrm r}({\boldsymbol z_t})\right ] =} \nonumber\\
    \label{eq32}
    &{\small {{\sum_{i=1}^{M}}}\left \{ \log \frac{\sigma^{\mathrm r}_{t,i}}{\sigma_{t,i}}+\frac{{\sigma_{t,i}}^2+(\mu_{t,i}-\mu^{\mathrm r}_{t,i})^2}{2{\sigma^{\mathrm r}_{t,i}}^2} -\frac{1}{2}\right \} }
\end{align}
\vspace{-0.4em}
 
The hyper-parameters ${{\boldsymbol\mu_t},\revisions{{\boldsymbol \Sigma}_{{\sf diag}, t}}}$\revisions{, where ${\sigma_{t,i}}^2$$=$${\Sigma}_{{\sf diag}, t, ii}$}, are calculated using an inference network $\boldsymbol \Phi^{\mathrm{infer}}({{{\boldsymbol w_0,\dots,\boldsymbol w_{t-1}}}})$, and ${{\boldsymbol \mu^{\mathrm r}_t},\revisions{{\boldsymbol \Sigma}^{\mathrm r}_{{\sf diag}, t}}}$\revisions{, where ${\sigma^{\mathrm r}_{t,i}}^2$$=$${\Sigma}^{\mathrm r}_{{\sf diag}, t, ii}$},  through an prior network $\boldsymbol \Phi^{\mathrm {prior}}({{{\boldsymbol w_0,\dots,\boldsymbol w_{t-1}}}})$ respectively.
{Both the inference network and prior network are regression networks.
The inputs of the two networks are history sequence $({{\boldsymbol w_0,\dots,\boldsymbol w_{t-1}}})$.
Each of the network produce two outputs including the mean and variance vectors of $q(\boldsymbol z_t)$ and $p_{\mathrm r}({\boldsymbol z_t})$ as follows.}
\vspace{-0.3em}
\begin{align}
    \label{eq33}
    {\small {[{\boldsymbol\mu_t}, \revisions{{\boldsymbol \Sigma}_{{\sf diag}, t}}]=\boldsymbol \Phi^{\mathrm{infer}}({{\boldsymbol w_0,\dots,\boldsymbol w_{t-1}}})}} \\ 
    \label{eq34}
    {\small {[{\boldsymbol \mu^{\mathrm r}_t}, \revisions{{\boldsymbol \Sigma}^{\mathrm r}_{{\sf diag}, t}}]=\boldsymbol \Phi^{\mathrm{prior}}({{\boldsymbol w_0,\dots,\boldsymbol w_{t-1}}})}}
\end{align}

\vspace{-0.3em}

{The input word sequence $({{\boldsymbol w_0,\dots,\boldsymbol w_{t-1}}})$ are approximated by the hidden state $\boldsymbol h_t$ in LSTM-RNNs, and the output of the Transformer feed-forward modules $\boldsymbol s_t^l$ as shown in Fig.\ref{fig:lstm} (d) and Fig.\ref{fig:trans} (d) respectively. 
In this work, both the inference networks and prior networks we use contain one single hidden layer for the Variational NNLMs.}
Both the inference and prior network modules are differentiable with respect to their internal parameters. 
Similar to the Bayesian and Gaussian Process NNLMs of Sec.\ref{sec:BNNLM} and \ref{sec:GPLM}, sampling the hidden output vector $\boldsymbol z_t$ from the {variational} distribution $q({\boldsymbol z_t})$ allows the marginal log-likelihood item $\mathcal L_1$ of Equation (\ref{eq30}) to be efficiently approximated. 
The maximization of the variational lower bound of Equation (\ref{eq30}) can then be integrated into the back-propagation algorithm during {variational} LSTM-RNN or Transformer LM training.

\section{System Implementation}
\label{sec:Imp}
In this section, a set of implementation details that affect the performance and efficiency of the Bayesian, Gaussian Process and Variational LSTM-RNN and Transformer LMs in Sec.\ref{sec:BLN} are discussed. 
\vspace{-0.4em}




\vspace{-0.5em}
\subsection{Choice of Prior Distributions}
\label{sec:prior}
When training all three Bayesian learning approaches based NNLMs, suitable choices of parameter prior distributions need to be set. 
In this paper, the prior distribution Gaussian means for various Bayesian learned LSTM-RNN and Transformer LMs are based on the parameter estimates of comparable deterministic, point estimated NNLMs that have converged in model training.
{Throughout this paper the prior Gaussian variances used for Bayesian learning based LSTM-RNN and Transformer LMs are empirically set as 1 and 1.0e-3 respectively.}
In addition, all the {other standard, non-Bayesian estimated parameters} in the Bayesian learning based LSTM-RNN and Transformer models are initialized using the parameters obtained from the comparable pre-trained standard models reaching half convergence. 
The combination of these two settings in practice was found to yield a good balance between convergence speed and performance\footnote{{Several methods to initialize the other standard non-Bayesian estimated model parameters include: 1) random weights; 2) the weights from both half-converged standard models; and 3) fully converged pre-trained standard models for Bayesian learning based NNLMs. 
Experimental results suggested that the model parameters initialized from the half-converged or fully converged pre-trained standard models produced similar performance, while both marginally better than using random weights.}}.
\vspace{-0.6em}

\begin{table*}[tb]
    \centering
    \scriptsize
    \captionsetup{font={small,it}}
    \caption{Perplexity (PPL) and WER(\%) of the baseline {\bf LSTM} and {\bf Transformer} LMs and their Bayesian variants on AMI {\bf dev} and {\bf eval} sets of {\bf ihm}, {\bf sdm1} and {\bf mdm8} conditions using different number of samples (1, 3 and 5) and {multiple seeds (1, 2 and 3)} after interpolation with 3-gram (3g) LM. 
    "FFN" represents the feed-forward network module in a {\bf Transformer} LM. 
    "$\dag$" and "$\ddagger$" denote statistically significant WER reductions were obtained over the baseline {\bf LSTM-RNN} (line 1) and {\bf Transformer} {(line 7)} LMs respectively.}
    \vspace{-0.5em}
    \resizebox{.87\textwidth}{!}{
    \begin{tabular}{c|l|cccc|c|cccc|c|cccc}
    \toprule
     \multirow{2}{*}{\bf ID} & \multirow{2}{*}{\bf LMs} &\multicolumn{4}{c|}{\bf Bayesian} & \bf PPL & \multicolumn{4}{c|}{\revisions{{\bf WER(\%)} of {\bf dev}}} & \bf PPL & \multicolumn{4}{c}{\revisions{{\bf WER(\%)} of {\bf eval}}} \\
     & & \multicolumn{1}{c}{\bf Layer} & \bf Position & \bf \#Sample & \bf Seed & \multicolumn{1}{c|}{{\bf (dev)}} & \bf ihm & \bf sdm1 & \bf mdm8 & \bf Avg. & \multicolumn{1}{c|}{{\bf (eval)}} & \bf ihm & \bf sdm1 & \bf mdm8 & \bf Avg. \\
    \hline
    \hline
     1 & LSTM+3g & \multicolumn{4}{c|}{Not Applied } & {60.4} & 16.4 & 30.0 & 27.7 & {24.7} & {59.6} & 16.2 & 34.8 & 30.9 & {27.3} \\
     \hline
     2 & \multirow{5}{*}{\shortstack{Bayesian LSTM\\+3g}} & 1 & \multirow{5}{*}{\shortstack{{Cell input}}} & 1 & 1 & {49.2} & 16.1$^\dag$ & 29.6$^\dag$ & 27.3$^\dag$ & {24.3}$^\dag$ & {50.5} & 15.9$^\dag$ & 34.3$^\dag$ & 30.5$^\dag$ & {26.9}$^\dag$ \\
     3 & & 1 & & 1 & 2 & {50.1} & {16.1}$^\dag$ & {29.6}$^\dag$ & {27.3}$^\dag$ & {24.3}$^\dag$ & {51.0} & {15.9}$^\dag$ & {34.3}$^\dag$ & {30.5}$^\dag$ & {26.9}$^\dag$ \\
     4 & & 1 & & 1 & 3 & {52.7} & {16.2} & {29.7} & {27.5} & {24.5} & {53.3} & {16.0} & {34.4} & {30.7} & {27.0} \\
     5 & & 1 & & \revision{3} & \revision{1} & \revision{48.3} & \revision{16.0}$^\dag$ & \revision{29.6}$^\dag$ & \revision{27.4}$^\dag$ & \revision{24.3}$^\dag$ & \revision{50.4} & \revision{15.9}$^\dag$ & \revision{34.2}$^\dag$ & \revision{30.4}$^\dag$ & \revision{26.8}$^\dag$ \\
     6 & & 1 & & \revision{5} & \revision{1} & \revision{49.5} & \revision{16.1}$^\dag$ & \revision{29.6}$^\dag$ & \revision{27.4}$^\dag$ & \revision{24.4}$^\dag$ & \revision{51.3} & \revision{15.8}$^\dag$ & \revision{34.3}$^\dag$ & \revision{30.4}$^\dag$ & \revision{26.8}$^\dag$ \\
     \hline
     \hline
     7 & Transformer+3g & \multicolumn{4}{c|}{Not Applied } & {46.9} & 16.1 & 29.9 & 27.5 & {24.5} & {48.1} & 15.9 & 34.5 & 30.5 & {27.0} \\
     \hline
     8 & \multirow{5}{*}{\shortstack{Bayesian Transformer\\+3g}} & 1 & \multirow{5}{*}{\shortstack{FFN}} & \revision{1} & \revision{1} & \revision{46.1} & \revision{16.0} & \revision{29.7}$^\ddagger$ & \revision{27.4} & \revision{24.4} & \revision{47.7} & \revision{15.8} & \revision{34.3}$^\ddagger$ & \revision{30.4} & \revision{26.8} \\
     9 & & 1 & & 1 & 2 & {46.2} & {16.0} & {29.7}$^\ddagger$ & {27.4} & {24.4} & {47.5} & {15.8} & {34.4} & {30.4} & {26.8} \\
     10 & & 1 & & 1 & 3 & {46.2} & {15.9} & {29.7}$^\ddagger$ & {27.4} & {24.4} & {47.4} & {15.8} & {34.3}$^\ddagger$ & {30.5} & {26.8} \\
     11 & & 1 & & \revision{3} & \revision{1} & \revision{47.0} & \revision{16.0} & \revision{29.7}$^\ddagger$ & \revision{27.4} & \revision{24.4} & \revision{48.2} & \revision{15.8} & \revision{34.4} & \revision{30.4} & \revision{26.9} \\
     12 & & 1 & & \revision{5} & \revision{1} & \revision{47.8} & \revision{16.0} & \revision{29.8} & \revision{27.4} & \revision{24.4} & \revision{49.3} & \revision{15.9} & \revision{34.4} & \revision{30.5} & \revision{26.9} \\
    \bottomrule
    \end{tabular}}
    \vspace{-1.85em}
\label{table:1}
\end{table*}

\vspace{-0.2em}
\subsection{Modeling Local Uncertainty}
\label{sec:uncertainty}
{When training various types of Bayesian estimated LSTM-RNN or Transformer LMs using variational inference, the evidence lower bounds in Equation (\ref{eq19}), (\ref{eq28}) and (\ref{eq30}) require Monte Carlo parameter samples to be drawn. 
These samples are drawn independently at each Bayesian estimated NNLM hidden layer. 
Assuming the number of samples drawn at each layer is $K$, the cost incurred in Bayesian inference grows exponentially with respect to the number of hidden layers $L$ as $K^{L}$.
Hence, directly performing Bayesian estimation using variational inference across too many NNLM hidden layers becomes computationally intractable.}

{One approach considered in this paper to address the above scalability issue is to use neural architecture search (NAS) techniques \cite{stanley2002evolving,kandasamy2018neural,zoph2016neural,liu2018darts,xie2018snas}.
They are used to automatically learn the most important network internal locations inside LSTM-RNNs and Transformer LMs that require Bayesian uncertainty modeling.
The form of NAS method considered in this paper is based on differentiable neural architecture search (DARTS) \cite{liu2018darts,xie2018snas,cai2018proxylessnas,hu2021neural,hu2020dsnas}. 
NAS using the DARTS method is performed over a super-network containing paths connecting either the \revision{point estimate}, or Bayesian estimated neural network structures to be considered.
The search requires the estimation of the weights that are assigned to each candidate neural architecture within such a super-network.
\revision{The search requires the estimation of the weights that are assigned to each point estimate based, or Bayesian estimated candidate neural architecture within such a super-network neural network LM.
The architecture weights and normal model parameters are jointly learned by minimizing the word sequence log-likelihood probability using the cross-entropy cost when training such a super-network language model.
The standard back-propagation algorithm is used in the training process.
When the super-network LM containing both the point estimate based, or Bayesian estimated architectures and their respective model parameters is trained to convergence, the optimal 1-best and top N-best architectures can be obtained by pruning lower weighted paths within the super-network that are considered less important},
akin to the approach used in our previous research on applying NAS to TDNNs \cite{hu2021neural}.}

{More specifically, the NNLM internal components to be examined during NAS include: a) the input, output, forget gates, cell inputs and hidden node activations (for Bayesian or GP LSTM-RNNs), as well as hidden output vectors (for variational LSTM-RNNs), as shown in Fig.\ref{fig:lstm}; 
and b) the multi-head self-attention and feedforward sub-layers within each Transformer LM block (for Bayesian or GP Transformers), as well as its hidden vector outputs (for variational Transformers), as shown in Fig.\ref{fig:trans}.}

The required DARTS super-network is constructed by building two parallel architecture paths for each of the above search locations within LSTM-RNN or Transformer LMs that respectively represent the use of either conventional \revisions{point estimate}, or Bayesian learning methods
of Sec.\ref{sec:BLN}.
For example, a fragment of such super-network designed for a cell input of an LSTM unit and its activation output is given below:
\vspace{-0.2em}
\begin{align}
    \label{eq35}
    &{\small \boldsymbol{\tilde c}=\frac{\exp{a^l_{\mathrm{\revisions{point}}}}}{\exp{a^l_{\mathrm{\revisions{point}}}}+\exp{a^l_{\mathrm{Bayes}}}}{\mathbf {tanh}}\left ( {{\boldsymbol \Theta}_{c,\mathrm{(\revisions{point})}} \left [ {{\boldsymbol x}_{t-1}}^{\top},{{\boldsymbol h}_{t-1}}^{\top},1 \right ]^\top} \right )}\nonumber\\
    &{\small +\frac{\exp{{{a^l_{\mathrm{Bayes}}}}}}{\exp{a^l_{\mathrm{\revisions{point}}}}+\exp{a^l_{\mathrm{Bayes}}}}{\mathbf {tanh}}\left ( {{\boldsymbol \Theta}_{c,\mathrm{(Bayes)}} \left [ {{\boldsymbol x}_{t-1}}^{\top},{{\boldsymbol h}_{t-1}}^{\top},1 \right ]^\top} \right )}
\end{align}
where $a^l_{\mathrm{\revisions{point}}}$ is the architecture weight indicating using \revision{point estimate} parameter estimation for the $l$-th {cell input}, and $a^l_{\mathrm{Bayes}}$ for using Bayesian parameter estimation.

\begin{figure}[!t]
\newcommand{\tabincell}[2]{\begin{tabular}{@{}#1@{}}#2\end{tabular}}
\setlength{\parskip}{0.0ex}
  \centering
  \vspace{-0.9em}
  \centerline{\includegraphics[width=6.0cm]{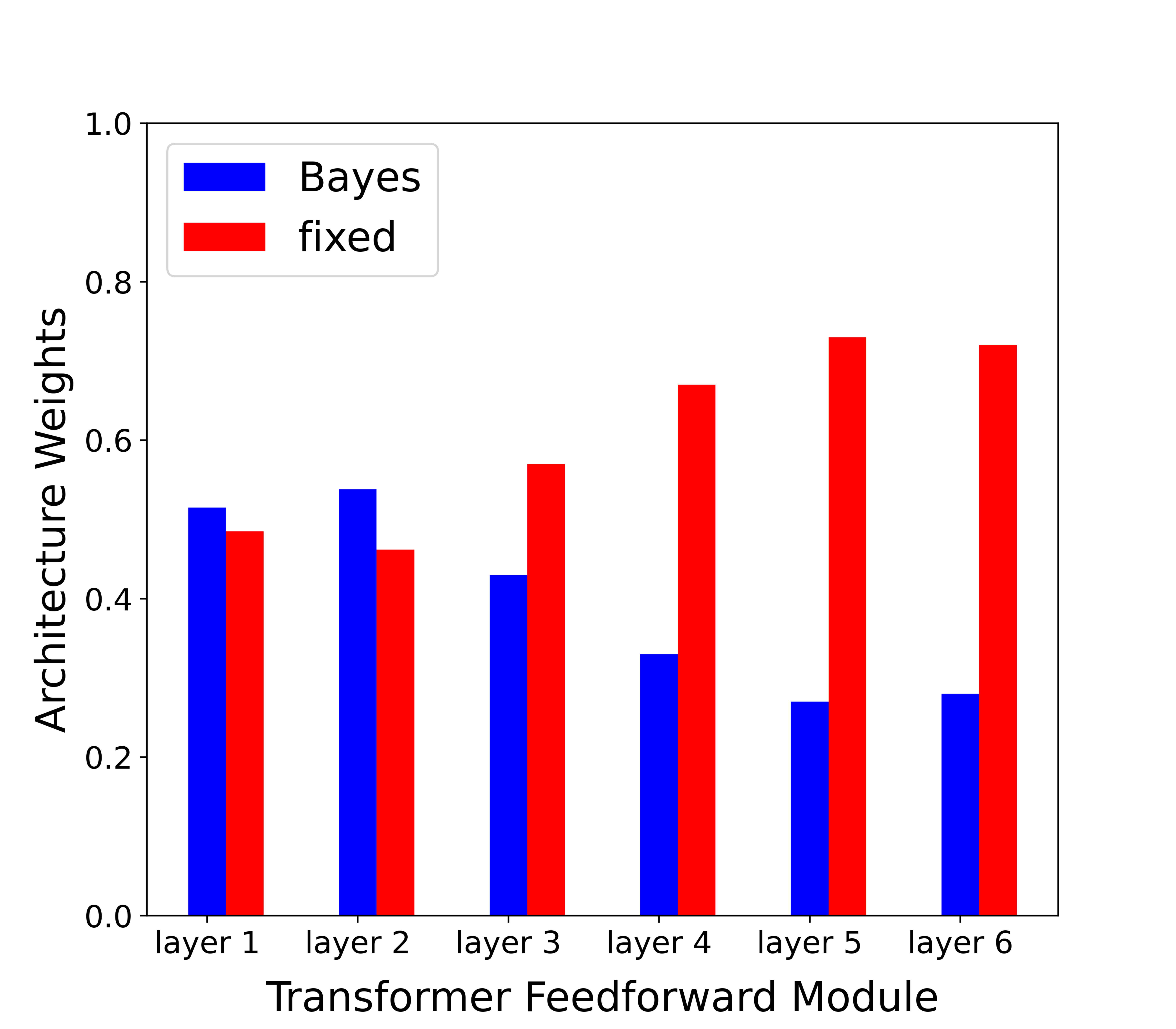}}
\captionsetup{font={small,it}}
\caption{{Examples of Transformer LM feed-forward module specific \revisions{point estimate} (in red) and Bayesian (in blue) architecture weights $a_{\mathrm {Bayes}}$, $a_{\mathrm {\revisions{point}}}$ extracted from the DARTS super-network trained on the AMI data. 
The bottom layers 1 and 2, where $a_{\mathrm {Bayes}}>a_{\mathrm {\revisions{point}}}$, were selected to be Bayesian estimated (shown in \revision{line 17 and 21}, Table \ref{table:4} with or without being interpolated with standard Transformer LMs).}}
\vspace{-1.5em}
\label{fig:nas}
\end{figure}

{Detailed experimental contrasts between using manual and NAS automatic selection of NNLM components to perform Bayesian estimation for LSTM-RNN and Transformer LMs are shown in Table \ref{table:3} (\revision{line} \revision{9-19 and 23} vs. \revision{line} \revision{20-22 and 24} for Bayesian LSTM-RNNs; and \revision{line} \revision{25-39 and 43} vs. \revision{line} \revision{40-42 and 44} for GP LSTM-RNNs) and Table \ref{table:4} (\revision{line} \revision{9-16 and 20} vs. \revision{line} \revision{17-19 and 21} for Bayesian Transformers; and \revision{line} \revision{22-27 and 31} vs. \revision{line} \revision{28-30 and 32} for GP Transformers) respectively.
A figure plotting the values of the six pairs of Transformer LM feed-forward module specific architecture weights $a_{\mathrm{Bayes}}$ (in blue) and $a_{\mathrm{\revisions{point}}}$ (in red) extracted from the DARTS super-network is shown in Fig.\ref{fig:nas}}\footnote{\revisions{The results suggested that the first and second layers in Transformer LM were selected to be Bayesian estimated. 
This decision is made based on the fact that the respective architecture weights associated with Bayesian estimation is larger than those of point estimation, $\alpha_{\mathrm{Bayes}}>\alpha_{\mathrm{point}}$, in the NAS super-network.
This can be intuitively interpreted as that the lower positioned, two bottom Transformer feedforward layers exhibit larger uncertainty than those positions at the top.
This is further confirmed by examining the median parameter signal-to-noise ratio (SNR) statistics \cite{braun2019parameter,blundell2015weight} (further discussed in section \ref{sec:exp_ami}) measured over the feedforward modules of a 6-layer Bayesian Transformer LM (line 16, Table \ref{table:4}).
The SNRs in Transformer layer 1-6 are 3.3, 3.1, 3.9, 4.7, 6.2, 7.0 respectively.
As expected, the medium parameter SNRs of the first two bottom layers are much lower than those of the higher layers. 
This trend indicates a larger parameter uncertainty in these two bottom layers that can benefit more from the use of Bayesian estimation.}}.
\vspace{-0.5em}

\vspace{-0.3em}
\subsection{Parameter Sampling}
\label{sec:sampling}
A second approach to further address the above scalability issue is to use a minimal number of parameter samples when approximating the marginal log-likelihood lower bounds in Equation (\ref{eq19}), (\ref{eq28}) and (\ref{eq30}) for the three types of Bayesian NNLMs of Sec.\ref{sec:BNNLM} to \ref{sec:VLM}. 
Within each individual component of NNLMs, for example, a {cell input} of an LSTM-RNN LM where Bayesian learning is applied and requires such parameter sampling, \revision{the model training cost can be further reduced when using a smaller number of samples.}

An ablation study on the performance of AMI meeting data trained Bayesian LSTM-RNN and Transformer LMs by varying the number of parameter samples drawn respectively at the first LSTM hidden layer’s {cell inputs} and the first Transformer module’s feed-forward sub-layer during model training {across multiple random seeds} is shown in Table \ref{table:1}. 
The experimental results in Table \ref{table:1} show that only a marginal difference in WER was observed by drawing more samples (three and five samples) in the {error forwarding}\footnote{“error forwarding” refers to the process of feed forwarding the training data through the neural network to compute the error loss for subsequent gradient backward propagation.} pass of Bayesian LSTM-RNN and Transformer LMs. 
Based on these findings,
only one sample is drawn to approximate the marginal log-likelihood in Equation (\ref{eq19}), (\ref{eq28}) and (\ref{eq30}) for all the Bayesian estimated NNLMs presented in this paper. 

During evaluation, the inference of Bayesian, Gaussian Process and Variational NNLMs are efficiently approximated by computing the expectation of the model parameters or the latent variables using their respective posterior distributions.
For example, during evaluation, the samples of activation parameters $\boldsymbol \Theta$ in Bayesian LSTM-RNN or Transformer LMs are approximated by taking the mean {${\boldsymbol \mu}$ in Equation (\ref{eq20})} of their latent distribution:
\begin{align}
    \label{eq36}
    {\small \int P({\boldsymbol W} |{\boldsymbol \Theta}) p({\boldsymbol \Theta} | {\cal D}) \mathrm d {\boldsymbol \Theta} \approx P(\boldsymbol W | \mathbb{E}[\boldsymbol \Theta| {\cal D}])}{=P(\boldsymbol W | \boldsymbol \mu)}
\end{align}
when predicting the probability of a test data sentence ${\boldsymbol W}$.




\begin{table}[tb]
    \centering
    \scriptsize
    \captionsetup{font={small,it}}
    \caption{Description of Bayesian, Gaussian Process and Variational NNLMs in terms of their respective uncertainty modeling, number of samples, number of free parameters and the speed ratios relative to the standard NNLMs in training and evaluation. The hidden layer input vector size is expressed as $a$, the number of hidden nodes as $b$, and the latent hidden output vector $\boldsymbol z$ is of $c$ dimensions. 
    }
    \vspace{-0.3em}
    \resizebox{.46\textwidth}{!}{
     \begin{tabular}{c|c|ccc|ccc|cc}
    \toprule
     \multirow{2}{*}{\bf ID} & \multirow{2}{*}{\shortstack{\bf LM}} & \multicolumn{3}{c|}{\bf Uncertainty} & \multicolumn{3}{c|}{\bf \#Parameter} & \multicolumn{2}{c}{\bf Speed Ratio} \\
     \cline{3-10}
     & & $\boldsymbol{\lambda}$ & $\boldsymbol{\theta}$ & $\boldsymbol{z}$ & $\boldsymbol{\lambda}$ & $\boldsymbol{\theta}$ & $\boldsymbol{z}$ & \bf Train & \bf Test \\
     \hline
     \hline
     1 & Standard & \xmark & \xmark & \xmark & 0 & ab & $0$ & 1.00 & 1.00 \\
     \hline
     2 & B-NNLMs & \xmark & \cmark & \xmark & 0 & 2ab & $0$ & 1.02 & 1.00 \\
     \hline
     3 & GP NNLMs & \cmark & \cmark & \xmark & 8b & 2ab & $0$ & 1.10 & 1.20 \\
     \hline
     4 & V-NNLMs & \xmark & \xmark & \cmark & $0$ & ab+cb & 4ac & 1.05 & 1.03 \\
    \bottomrule
    \end{tabular}}
    \vspace{-1.7em}
    \label{table:2}
\end{table}

\subsection{System Description} 
\label{sec:description}
Following the above implementation details, the description of a set of Bayesian, Gaussian Process and Variational NNLMs in terms of their respective forms of uncertainty modeling, number of free parameters and the speed ratios relative to the standard NNLMs in training and evaluation is presented in Table \ref{table:2}. 
In addition, when using the above efficient sampling during inference for model training (as low as one sample drawn) and evaluation, the Bayesian, Gaussian Process and Variational NNLMs only require a moderate increase in system training time of approximately 5\%-10\% over the standard baseline NNLMs during training, while their computational complexity is comparable to that of standard NNLMs during evaluation. 

\begin{table*}[htb]
    \centering
    \scriptsize
    \captionsetup{font={small,it}}
    \caption{Perplexity and WER\% of the baseline 3-gram (3g), {\bf LSTM-RNN} LMs with \revision{standard point estimate}, {and L1, or L2 regularized or MAP estimated {\bf LSTM-RNN} LMs}, and Bayesian, GP and variational {\bf LSTM-RNN} LMs with various forms of uncertainty modeling on the AMI {\bf dev} and {\bf eval} sets of {\bf ihm}, {\bf sdm1} and {\bf mdm8} conditions. 
    \revision{"Design" (also in Table \ref{table:4} and \ref{table:6}) denotes whether the network internal positions to be Bayesian estimated are selected manually (Manual), or automatically using NAS on the listed search space.}
    "+3g" and "+LSTM" denote the results of further interpolation with the baseline 3-gram and {\bf LSTM-RNN} LMs. "†" denotes statistically significant WER reductions were obtained over the {\bf LSTM-RNN} baseline (line 2).
    }
    \vspace{-0.5em}
    \resizebox{.98\textwidth}{!}{
    \begin{tabular}{c|l|c|cc|ccc|c|cccc|c|cccc}
    \toprule
     \multirow{2}{*}{\bf ID} & \multirow{2}{*}{\bf LMs} &\multicolumn{3}{c|}{\bf Bayesian} & \multicolumn{3}{c|}{\bf Uncertainty} & \bf PPL & \multicolumn{4}{c|}{\revisions{{\bf WER(\%)} of {\bf dev}}} & \bf PPL & \multicolumn{4}{c}{\revisions{{\bf WER(\%)} of {\bf eval}}} \\
     & & \multicolumn{1}{c}{\bf Design} & \bf Block & \bf Position & \scalebox{0.9}{$\boldsymbol{\lambda}$} & \scalebox{0.9}{$\boldsymbol{\theta}$} & \scalebox{0.9}{$\boldsymbol{z}$} & \multicolumn{1}{c|}{{\bf (dev)}} & \bf ihm & \bf sdm1 & \bf mdm8 & \bf Avg. & \multicolumn{1}{c|}{{\bf (eval)}} & \bf ihm & \bf sdm1 & \bf mdm8 & \bf Avg. \\
    \hline
    \hline
     1 & 3-gram & \multicolumn{3}{c|}{\multirow{8}{*}{\shortstack{\scalebox{0.9}{Not Applied}}}} & \multirow{8}{*}{\shortstack{\scalebox{0.9}{\xmark}}} & \multirow{8}{*}{\shortstack{\scalebox{0.9}{\xmark}}} & \multirow{8}{*}{\shortstack{\scalebox{0.9}{\xmark}}} & {81.9} & 18.1 & 31.8 & 29.5 & {26.5} & {93.6} & 18.4 & 36.4 & 32.5 & {29.1} \\
     2 & \scalebox{0.9}{LSTM +3g} & \multicolumn{3}{c|}{ } & & & & {60.4} & 16.4 & 30.0 & 27.7 & {24.7} & {59.6} & 16.2 & 34.7 & 30.9 & {27.3} \\
     \cline{1-2}\cline{9-18}
     3 & \scalebox{0.9}{{LSTM(L1) +3g}} & \multicolumn{3}{c|}{ } & & & & {54.5} & {16.3} & {29.8} & {27.6} & {24.6} & {55.6} & {16.1} & {34.5} & {30.7} & {27.1} \\
     4 & \scalebox{0.9}{{\ \ \revision{+LSTM}}} & \multicolumn{3}{c|}{ } & & & & \revision{53.2} & \revision{16.2} & \revision{29.7} & \revision{27.5} & \revision{24.5} & \revision{54.2} & \revision{16.0} & \revision{34.4} & \revision{30.6} & \revision{27.0} \\
     \cline{2-2}\cline{9-18}
     5 & \scalebox{0.9}{{LSTM(L2) +3g}} & \multicolumn{3}{c|}{ } & & & & {53.4} & {16.3} & {29.8} & {27.5} & {24.5} & {54.7} & {16.0} & {34.6} & {30.7} & {27.1} \\
     6 & \scalebox{0.9}{{\ \ \revision{+LSTM}}} & \multicolumn{3}{c|}{ } & & & & \revision{52.3} & \revision{16.2} & \revision{29.7} & \revision{27.4} & \revision{24.4} & \revision{53.1} & \revision{15.9} & \revision{34.4} & \revision{30.5} & \revision{26.9} \\
     \cline{2-2}\cline{9-18}
     7 & \scalebox{0.9}{{LSTM(MAP) +3g}} & \multicolumn{3}{c|}{ } & & & & {52.5} & {16.2} & {29.7} & {27.5} & {24.5} & {53.7} & {16.0} & {34.5} & {30.6} & {27.0} \\
     8 & \scalebox{0.9}{{\ \ \revision{+LSTM}}} & \multicolumn{3}{c|}{ } & & & & \revision{51.4} & \revision{16.1} & \revision{29.6} & \revision{27.4} & \revision{24.4} & \revision{52.4} & \revision{15.9} & \revision{34.4} & \revision{30.5} & \revision{26.9} \\
    \hline
    \hline
     9 & \multirow{14}{*}{\shortstack{\scalebox{0.9}{Bayesian}\\\scalebox{0.9}{LSTM}\\+3g}} & \multirow{11}{*}{Manual} & 1-2 & \multirow{3}{*}{\shortstack{\scalebox{0.9}{All gates}}} & \multirow{3}{*}{\shortstack{\scalebox{0.9}{\xmark}}} & \multirow{3}{*}{\shortstack{\scalebox{0.9}{\cmark}}} & \multirow{3}{*}{\shortstack{\scalebox{0.9}{\xmark}}} & {52.1} & 16.2$^\dag$ & 29.6$^\dag$ & 27.4$^\dag$ & {24.4}$^\dag$ & {53.3} & 16.0$^\dag$ & 34.5$^\dag$ & 30.6$^\dag$ & {27.0}$^\dag$ \\
     10 & & & 1 & & & & & {48.2} & 16.0$^\dag$ & 29.5$^\dag$ & 27.3$^\dag$ & {24.3}$^\dag$ & {49.5} & 15.8$^\dag$ & 34.3$^\dag$ & 30.4$^\dag$ & {26.8}$^\dag$ \\
     11 & & & 2 & & & & & {46.9} & 16.0$^\dag$ & 29.5$^\dag$ & \bf27.2$^\dag$ & {24.2}$^\dag$ & {48.1} & 15.7$^\dag$ & 34.2$^\dag$ & 30.3$^\dag$ & {26.7}$^\dag$ \\
     \cline{4-18}
     12 & & & 1 & \scalebox{0.9}{Input gate}\scalebox{0.8}{(IG)} & \multirow{4}{*}{\shortstack{\scalebox{0.9}{\xmark}}} & \multirow{4}{*}{\shortstack{\scalebox{0.9}{\cmark}}} & \multirow{4}{*}{\shortstack{\scalebox{0.9}{\xmark}}} & {52.1} & 16.2$^\dag$ & 29.6$^\dag$ & 27.5$^\dag$ & {24.4}$^\dag$ & {54.3} & 15.9$^\dag$ & 34.4$^\dag$ & 30.6$^\dag$ & {27.0}$^\dag$ \\
     13 & & & 1 & \scalebox{0.9}{Forget gate}\scalebox{0.8}{(FG)} & & & & {52.2} & 16.2$^\dag$ & 29.7$^\dag$ & 27.4$^\dag$ & {24.4}$^\dag$ & {53.9} & 16.0$^\dag$ & 34.4$^\dag$ & 30.6$^\dag$ & {27.0}$^\dag$ \\
     14 & & & 1 & \scalebox{0.9}{{Cell input}}\scalebox{0.8}{(CI)} & & & & {49.2} & 16.1$^\dag$ & 29.6$^\dag$ & 27.3$^\dag$ & {24.3}$^\dag$ & {50.5} & 15.9$^\dag$ & 34.3$^\dag$ & 30.5$^\dag$ & {26.9}$^\dag$ \\
     15 & & & 1 & \scalebox{0.9}{Output gate}\scalebox{0.8}{(OG)} & & & & {52.5} & 16.2$^\dag$ & 29.7$^\dag$ & 27.5$^\dag$ & {24.5}$^\dag$ & {53.0} & 16.0$^\dag$ & 34.4$^\dag$ & 30.6$^\dag$ & {27.0}$^\dag$ \\
     \cline{4-18}
     16 & & & 2 & \scalebox{0.9}{Input gate}\scalebox{0.8}{(IG)} & \multirow{4}{*}{\shortstack{\scalebox{0.9}{\xmark}}} & \multirow{4}{*}{\shortstack{\scalebox{0.9}{\cmark}}} & \multirow{4}{*}{\shortstack{\scalebox{0.9}{\xmark}}} & {51.4} & 16.2$^\dag$ & 29.6$^\dag$ & 27.3$^\dag$ & {24.4}$^\dag$ & {52.5} & 15.9$^\dag$ & 34.5$^\dag$ & 30.6$^\dag$ & {27.0}$^\dag$ \\
     17 & & & 2 & \scalebox{0.9}{Forget gate}\scalebox{0.8}{(FG)} & & & & {51.4} & 16.2$^\dag$ & 29.6$^\dag$ & 27.4$^\dag$ & {24.4}$^\dag$ & {52.4} & 15.9$^\dag$ & 34.4$^\dag$ & 30.6$^\dag$ & {27.0}$^\dag$ \\
     18 & & & 2 & \scalebox{0.9}{{Cell input}}\scalebox{0.8}{(CI)} & & & & {47.6} & 16.0$^\dag$ & 29.6$^\dag$ & 27.3$^\dag$ & {24.3}$^\dag$ & {48.0} & 15.7$^\dag$ & 34.2$^\dag$ & 30.3$^\dag$ & {26.7}$^\dag$ \\
     19 & & & 2 & \scalebox{0.9}{Output gate}\scalebox{0.8}{(OG)} & & & & {51.2} & 16.2$^\dag$ & 29.6$^\dag$ & 27.4$^\dag$ & {24.4}$^\dag$ & {52.0} & 15.9$^\dag$ & 34.4$^\dag$ & 30.6$^\dag$ & {27.0}$^\dag$ \\
     \cline{3-18}
     20 & & \multirow{3}{*}{\shortstack{\scalebox{0.9}{NAS} \scalebox{0.8}{(Search Space:}\\\scalebox{0.9}{Layer 1,2}\\\scalebox{0.9}{IG,FG,{CI},OG)}}} & 1-2 (top1) & \multirow{3}{*}{\shortstack{\scalebox{0.9}{{Cell input}}}} & \multirow{3}{*}{\shortstack{\scalebox{0.9}{\xmark}}} & \multirow{3}{*}{\shortstack{\scalebox{0.9}{\cmark}}} & \multirow{3}{*}{\shortstack{\scalebox{0.9}{\xmark}}} & {46.0} & \bf15.9$^\dag$ & 29.5$^\dag$ & 27.2$^\dag$ & {24.2}$^\dag$ & {47.1} & 15.6$^\dag$ & 34.2$^\dag$ & \bf30.2$^\dag$ & {26.7}$^\dag$ \\
     21 & & & 1 (top2) & & & & & {49.2} & 16.1$^\dag$ & 29.6$^\dag$ & 27.3$^\dag$ & {24.3}$^\dag$ & {50.5} & 15.9$^\dag$ & 34.3$^\dag$ & 30.5$^\dag$ & {26.9}$^\dag$ \\
     22 & & & 2 (top3) & & & & & {47.6} & 16.0$^\dag$ & 29.6$^\dag$ & 27.3$^\dag$ & {24.3}$^\dag$ & {48.0} & 15.7$^\dag$ & 34.2$^\dag$ & 30.3$^\dag$ & {26.7}$^\dag$ \\
     \hline
     23 & \multirow{2}{*}{\shortstack{\scalebox{0.9}{\ \ +LSTM}}} & \multirow{1}{*}{Manual} & 2 & \scalebox{0.9}{All gates} & \multirow{1}{*}{\shortstack{\scalebox{0.9}{\xmark}}} & \multirow{1}{*}{\shortstack{\scalebox{0.9}{\cmark}}} & \multirow{1}{*}{\shortstack{\scalebox{0.9}{\xmark}}} & \revision{47.4} & \bf15.9$^\dag$ & \bf29.4$^\dag$ & 27.2$^\dag$ & {24.2}$^\dag$ & \revision{48.3} & \bf15.6$^\dag$ & \bf34.1$^\dag$ & \bf{30.2}$^\dag$ & {\bf26.6}$^\dag$ \\
     \cline{3-18}
     24 & & \multirow{1}{*}{\shortstack{NAS}} & 1-2 & \scalebox{0.9}{{Cell input}} & \multirow{1}{*}{\shortstack{\scalebox{0.9}{\xmark}}} & \multirow{1}{*}{\shortstack{\scalebox{0.9}{\cmark}}} & \multirow{1}{*}{\shortstack{\scalebox{0.9}{\xmark}}} & \revision{47.1} & \bf15.9$^\dag$ & \bf29.4$^\dag$ & \bf27.1$^\dag$ & {\bf24.1}$^\dag$ & \revision{47.3} & {\bf15.6}$^\dag$ & \bf34.1$^\dag$ & \bf{30.2}$^\dag$ & {\bf26.6}$^\dag$ \\
     \hline
     \hline
     25 & \multirow{21}{*}{\shortstack{\scalebox{0.9}{GP LSTM}\\\scalebox{0.9}{+3g}}} & \multirow{18}{*}{Manual} & 1 & \scalebox{0.9}{Input gate}\scalebox{0.8}{({IG})} & \multirow{7}{*}{\shortstack{\scalebox{0.9}{\cmark}}} & \multirow{7}{*}{\shortstack{\scalebox{0.9}{\cmark}}} & \multirow{7}{*}{\shortstack{\scalebox{0.9}{\xmark}}} & {58.3} & 16.4 & 30.0 & 27.7 & {24.7} & {57.6} & 16.3 & 34.8 & 30.9 & {27.3} \\
     26 & & & 1 & \scalebox{0.9}{Forget gate}\scalebox{0.8}{({FG})} & & & & {54.3} & 16.1$^\dag$ & 29.9 & 27.5$^\dag$ & {24.5} & {55.2} & 16.1 & 34.6 & 30.8 & {27.2} \\
     27 & & & 1 & \scalebox{0.9}{{Cell input}}\scalebox{0.8}{({CI})} & & & & {52.4} & 16.0$^\dag$ & 29.7$^\dag$ & 27.4$^\dag$ & {24.4}$^\dag$ & {53.8} & 16.0 & 34.5$^\dag$ & 30.4$^\dag$ & {27.0}$^\dag$ \\
     28 & & & 1 & \scalebox{0.9}{Output gate}\scalebox{0.8}{({OG})} & & & & {50.1} & 16.1$^\dag$ & 29.7$^\dag$ & 27.5$^\dag$ & {24.4}$^\dag$ & {51.5} & 16.0 & 34.5$^\dag$ & 30.5$^\dag$ & {27.0}$^\dag$ \\
     29 & & & 1 & \scalebox{0.9}{c-Gate}\scalebox{0.8}{({cG})} & & & & {54.4} & 16.2$^\dag$ & 29.9 & 27.5 & {24.5} & {55.7} & 16.1 & 34.6 & 30.7 & {27.1} \\
     30 & & & 1 & \scalebox{0.9}{h-Gate}\scalebox{0.8}{({hG})} & & & & {51.2} & 16.1$^\dag$ & 29.6$^\dag$ & 27.4$^\dag$ & {24.4}$^\dag$ & {52.5} & 16.0 & 34.4$^\dag$ & 30.6$^\dag$ & {27.0}$^\dag$ \\
     31 & & & 1 & \scalebox{0.9}{i-Gate}\scalebox{0.8}{(iG)} & & & & {56.7} & 16.3 & 29.9 & 27.7 & {24.6} & {56.9} & 16.2 & 34.7 & 30.8 & {27.2} \\
     \cline{4-18}
     32 & & & 2 & \scalebox{0.9}{Input gate}\scalebox{0.8}{(IG)} & \multirow{7}{*}{\shortstack{\scalebox{0.9}{\cmark}}} & \multirow{7}{*}{\shortstack{\scalebox{0.9}{\cmark}}} & \multirow{7}{*}{\shortstack{\scalebox{0.9}{\xmark}}} & {57.3} & 16.3 & 30.0 & 27.8 & {24.7} & {58.0} & 16.3 & 34.8 & 31.0 & {27.4} \\
     33 & & & 2 & \scalebox{0.9}{Forget gate}\scalebox{0.8}{(FG)} & & & & {58.6} & 16.4 & 30.1 & 27.8 & {24.8} & {57.5} & 16.3 & 34.9 & 31.2 & {27.5} \\
     34 & & & 2 & \scalebox{0.9}{{Cell input}}\scalebox{0.8}{(CI)} & & & & {50.4} & 16.0$^\dag$ & 29.6$^\dag$ & 27.4$^\dag$ & {24.3}$^\dag$ & {51.2} & 15.9$^\dag$ & 34.4$^\dag$ & 30.4$^\dag$ & {26.9}$^\dag$ \\
     35 & & & 2 & \scalebox{0.9}{Output gate}\scalebox{0.8}{(OG)} & & & & {53.1} & 16.1$^\dag$ & 29.8 & 27.5$^\dag$ & {24.5}$^\dag$ & {54.5} & 16.0 & 34.5$^\dag$ & 30.6$^\dag$ & {27.0}$^\dag$ \\
     36 & & & 2 & \scalebox{0.9}{c-Gate}\scalebox{0.8}{(cG)} & & & & {65.5} & 16.6 & 30.2 & 27.9 & {24.9} & {64.0} & 16.5 & 34.8 & 31.2 & {27.5} \\
     37 & & & 2 & \scalebox{0.9}{h-Gate}\scalebox{0.8}{(hG)} & & & & {53.3} & 16.2$^\dag$ & 29.7$^\dag$ & 27.5$^\dag$ & {27.5}$^\dag$ & {54.6} & 16.0 & 34.7 & 30.7$^\dag$ & {27.1} \\
     38 & & & 2 & \scalebox{0.9}{i-Gate}\scalebox{0.8}{(iG)} & & & & {55.7} & 16.3 & 29.8 & 27.7 & {27.6} & {58.3} & 16.1 & 34.6 & 30.7 & {27.1} \\
    \cline{4-18}
     39 & & & 1-2 & \scalebox{0.9}{{Cell input}}\scalebox{0.8}{(CI)} & \multirow{1}{*}{\shortstack{\scalebox{0.9}{\cmark}}} & \multirow{1}{*}{\shortstack{\scalebox{0.9}{\cmark}}} & \multirow{1}{*}{\shortstack{\scalebox{0.9}{\xmark}}} & {48.4} & 16.1$^\dag$ & 29.6$^\dag$ & 27.2$^\dag$ & {24.3}$^\dag$ & {48.2} & 15.8$^\dag$ & 34.3$^\dag$ & 30.5$^\dag$ & {26.9}$^\dag$ \\
    \cline{3-18}
     40 & & \multirow{3}{*}{\shortstack{\scalebox{0.9}{NAS} \scalebox{0.8}{(Search Space:}\\\scalebox{0.9}{Layer 1,2}\\\scalebox{0.9}{{CI},OG,hG)}}} & 1-2 (top1) & \multirow{3}{*}{\shortstack{\scalebox{0.9}{h-Gate}}} & \multirow{3}{*}{\shortstack{\scalebox{0.9}{\cmark}}} & \multirow{3}{*}{\shortstack{\scalebox{0.9}{\cmark}}} & \multirow{3}{*}{\shortstack{\scalebox{0.9}{\xmark}}} & {48.1} & 16.0$^\dag$ & 29.5$^\dag$ & 27.4$^\dag$ & {24.3}$^\dag$ & {48.9} & 15.8$^\dag$ & 34.3$^\dag$ & 30.5$^\dag$ & {26.9}$^\dag$ \\
     41 & & & 1 (top2) & & & & & {51.2} & 16.1$^\dag$ & 29.6$^\dag$ & 27.4$^\dag$ & {24.4}$^\dag$ & {52.6} & 16.0 & 34.4$^\dag$ & 30.6$^\dag$ & {27.0}$^\dag$ \\
     42 & & & 2 (top3) & & & & & {53.3} & 16.2$^\dag$ & 29.7$^\dag$ & {27.5}$^\dag$ & {27.5}$^\dag$ & {55.6} & 16.0 & 34.7 & 30.7$^\dag$ & {27.1}$^\dag$ \\
     \hline
     43 & \multirow{2}{*}{\shortstack{\scalebox{0.9}{\ \ +LSTM}}} & \multirow{1}{*}{Manual} & 1-2 & \scalebox{0.9}{{Cell input}} & \multirow{1}{*}{\shortstack{\scalebox{0.9}{\cmark}}} & \multirow{1}{*}{\shortstack{\scalebox{0.9}{\cmark}}} & \multirow{1}{*}{\shortstack{\scalebox{0.9}{\xmark}}} & \revision{48.2} & 15.9$^\dag$ & 29.5$^\dag$ & \bf27.2$^\dag$ & {24.2}$^\dag$ & \revision{48.6} & 15.7$^\dag$ & \bf34.2$^\dag$ & 30.5$^\dag$ & {26.8}$^\dag$ \\
    \cline{3-18}
     44 & & \multirow{1}{*}{\shortstack{NAS}} & 1-2 & \scalebox{0.9}{h-Gate} & \multirow{1}{*}{\shortstack{\scalebox{0.9}{\cmark}}} & \multirow{1}{*}{\shortstack{\scalebox{0.9}{\cmark}}} & \multirow{1}{*}{\shortstack{\scalebox{0.9}{\xmark}}} & \revision{47.6} & \bf15.8$^\dag$ & \bf29.4$^\dag$ & \bf27.2$^\dag$ & {\bf 24.1}$^\dag$ & \revision{49.2} & {\bf15.6}$^\dag$ & \bf34.2$^\dag$ & \bf30.3$^\dag$ & {\bf 26.7}$^\dag$ \\
     \hline
     \hline
     45 & \multirow{3}{*}{\shortstack{\scalebox{0.9}{Variational} \scalebox{0.9}{LSTM}\\\scalebox{0.9}{+3g}}} & \multirow{3}{*}{Manual} & 1 & \multirow{3}{*}{\shortstack{\scalebox{0.9}{Hidden output}}} & \multirow{3}{*}{\shortstack{\scalebox{0.9}{\xmark}}} & \multirow{3}{*}{\shortstack{\scalebox{0.9}{\xmark}}} & \multirow{3}{*}{\shortstack{\scalebox{0.9}{\cmark}}} & {53.6} & 16.2 & 29.7 & 27.6 & {24.5} & {54.7} & 16.0 & 34.5 & 30.7 & {27.1} \\
     46 & & & 2 & & & & & {53.6} & 16.2 & 29.8 & 27.6 & {24.5} & {55.0} & 16.0 & 34.5 & 30.7 & {27.1} \\
     47 & & & 1-2 & & & & & {52.7} & 16.2 & 29.7 & 27.5 & {24.4} & {53.8} & 15.9 & 34.4 & 30.6 & {27.0} \\
     \hline
     48 & \multirow{1}{*}{\shortstack{\scalebox{0.9}{\ \ +LSTM}}} & \multirow{1}{*}{Manual} & 1-2 & \multirow{1}{*}{\shortstack{\scalebox{0.9}{Hidden output}}} & \multirow{1}{*}{\shortstack{\scalebox{0.9}{\xmark}}} & \multirow{1}{*}{\shortstack{\scalebox{0.9}{\xmark}}} & \multirow{1}{*}{\shortstack{\scalebox{0.9}{\cmark}}} & \revision{51.4} & 16.0$^\dag$ & 29.5$^\dag$ & 27.4$^\dag$ & {24.3}$^\dag$ & \revision{52.4} & 15.7$^\dag$ & 34.3$^\dag$ & 30.5$^\dag$ & {26.8}$^\dag$ \\
     \hline
    \bottomrule
    \end{tabular}}
    \vspace{-2.2em}
    \label{table:3}
\end{table*}

\begin{table*}[htb]
    \centering
    \scriptsize
    \captionsetup{font={small,it}}
    \caption{Perplexity and WER\% of the baseline 3-gram (3g), {\bf Transformer} LMs with \revision{standard point estimate}, {and L1, or L2 regularized or MAP estimated {\bf Transformer} LMs}, and Bayesian, GP and variational {\bf Transformer} LMs with various forms of uncertainty modeling on the AMI {\bf dev} and {\bf eval} sets of {\bf ihm}, {\bf sdm1} and {\bf mdm8} conditions. 
    "FFN", "MHA" and "EMB" denote the feed-forward, multi-head self-attention and embedding layers respectively.
    "+3g" and "+Transformer" denote the results of further interpolation with the baseline 3-gram and {\bf Transformer} LMs. 
    "†" denotes statistically significant WER reductions were obtained over the {\bf Transformer} baseline (line 2).
    }
    \vspace{-0.5em}
    \resizebox{.98\textwidth}{!}{
    \begin{tabular}{c|l|c|cc|ccc|c|cccc|c|cccc}
    \toprule
     \multirow{2}{*}{\bf ID} & \multirow{2}{*}{\bf LMs} &\multicolumn{3}{c|}{\bf Bayesian} & \multicolumn{3}{c|}{\bf Uncertainty} & \bf PPL & \multicolumn{4}{c|}{\revisions{{\bf WER(\%)} of {\bf dev}}} & \bf PPL & \multicolumn{4}{c}{\revisions{{\bf WER(\%)} of {\bf eval}}} \\
     & & \multicolumn{1}{c}{\bf Design} & \bf Block & \bf Position & \scalebox{0.9}{$\boldsymbol{\lambda}$} & \scalebox{0.9}{$\boldsymbol{\theta}$} & \scalebox{0.9}{$\boldsymbol{z}$} & \multicolumn{1}{c|}{{\bf (dev)}} & \bf ihm & \bf sdm1 & \bf mdm8 & \bf Avg. & \multicolumn{1}{c|}{{\bf (eval)}} & \bf ihm & \bf sdm1 & \bf mdm8 & \bf Avg. \\
    \hline
    \hline
     1 & 3-gram  & \multicolumn{3}{c|}{\multirow{8}{*}{\shortstack{\scalebox{0.9}{Not Applied}}}} & \multirow{8}{*}{\shortstack{\scalebox{0.9}{\xmark}}} & \multirow{8}{*}{\shortstack{\scalebox{0.9}{\xmark}}} & \multirow{8}{*}{\shortstack{\scalebox{0.9}{\xmark}}} & {81.9} & 18.1 & 31.8 & 29.5 & {26.5} & {93.6} & 18.4 & 36.4 & 32.5 & {29.1} \\
     2 & \scalebox{0.9}{Transformer +3g} & \multicolumn{3}{c|}{ } & & & & {46.9} & 16.1 & 29.9 & 27.5 & {24.5} & {48.4} & 15.9 & 34.5 & 30.5 & {27.0} \\
     \cline{1-2}\cline{9-18}
     3 & \scalebox{0.9}{{Transformer(L1) +3g}} & \multicolumn{3}{c|}{ } & & & & {46.8} & {16.1} & {29.9} & {27.5} & {24.5} & {48.3} & {15.9} & {34.5} & {30.6} & {27.0} \\
     4 & \scalebox{0.9}{{\ \ \revision{+Transformer}}} & \multicolumn{3}{c|}{ } & & & & \revision{46.8} & \revision{16.0} & \revision{29.8} & \revision{27.4} & \revision{24.4} & \revision{48.1} & \revision{15.8} & \revision{34.5} & \revision{30.5} & \revision{26.9} \\
     \cline{2-2}\cline{9-18}
     5 & \scalebox{0.9}{{Transformer(L2) +3g}} & \multicolumn{3}{c|}{ } & & & & {46.8} & {16.1} & {29.9} & {27.5} & {24.5} & {48.2} & {15.9} & {34.5} & {30.4} & {26.9} \\
     6 & \scalebox{0.9}{{\ \ \revision{+Transformer}}} & \multicolumn{3}{c|}{ } & & & & \revision{46.7} & \revision{16.0} & \revision{29.8} & \revision{27.4} & \revision{24.4} & \revision{48.1} & \revision{15.8} & \revision{34.4} & \revision{30.4} & \revision{26.9} \\
     \cline{2-2}\cline{9-18}
     7 & \scalebox{0.9}{{Transformer(MAP) +3g}} & \multicolumn{3}{c|}{ } & & & & {47.1} & {16.1} & {29.9} & {27.5} & {24.5} & {48.3} & {15.9} & {34.5} & {30.5} & {27.0} \\
     8 & \scalebox{0.9}{{\ \ \revision{+Transformer}}} & \multicolumn{3}{c|}{ } & & & & \revision{46.8} & \revision{16.0} & \revision{29.8} & \revision{27.4} & \revision{24.4} & \revision{48.1} & \revision{15.8} & \revision{34.4} & \revision{30.5} & \revision{26.9} \\
     \hline
     \hline
     9 & \multirow{11}{*}{\shortstack{\scalebox{0.9}{Bayesian}\\ \scalebox{0.9}{Transformer}\\\scalebox{0.9}{+3g}}} &\multirow{8}{*}{Manual} & - & \scalebox{0.9}{EMB} & \multirow{3}{*}{\shortstack{\scalebox{0.9}{\xmark}}} & \multirow{3}{*}{\shortstack{\scalebox{0.9}{\cmark}}} & \multirow{3}{*}{\shortstack{\scalebox{0.9}{\xmark}}} & {48.1} & 16.1 & 29.8 & 27.5 & {24.5} & {49.4} & 15.9 & 34.5 & 30.5 & {27.0} \\
     10 & && 1 & \scalebox{0.9}{MHA} & & & & {47.0} & 16.0 & 29.8 & 27.4 & {24.4} & {48.2} & 15.8 & 34.4 & 30.5 & {26.9} \\
     11 & && 1 & \scalebox{0.9}{FFN} & & & & {46.6} & 16.0 & 29.7$^\dag$ & 27.4 & {24.4} & {47.3} & 15.8 & 34.3$^\dag$ & 30.4 & {26.8} \\
     \cline{4-18}
     12 & & & 1-2 & \multirow{5}{*}{\scalebox{0.9}{FFN}} & \multirow{5}{*}{\shortstack{\scalebox{0.9}{\xmark}}} & \multirow{5}{*}{\shortstack{\scalebox{0.9}{\cmark}}} & \multirow{5}{*}{\shortstack{\scalebox{0.9}{\xmark}}} & {47.1} & 16.0 & 29.7 & 27.4 & {24.4} & {48.0} & 15.8 & 34.4 & 30.5 & {26.9} \\
     13 & & & 1-3 & & & & & {47.2} & 16.0 & 29.8 & 27.5 & {24.4} & {48.3} & 15.8 & 34.5 & 30.5 & {26.9} \\
     14 & & & 1-4 & & & & & {47.8} & 16.0 & 29.8 & 27.5 & {24.4} & {49.2} & 15.9 & 34.5 & 30.7 & {27.0} \\
     15 & & & 1-5 & & & & & {47.9} & 16.1 & 29.9 & 27.7 & {24.6} & {49.6} & 15.9 & 34.4 & 30.7 & {27.0} \\
     16 & & & 1-6 & & & & & {49.2} & 16.1 & 30.0 & 27.7 & {24.6} & {51.5} & 16.0 & 34.5 & 30.7 & {27.1} \\
     \cline{3-18}
     17 & & \multirow{3}{*}{\shortstack{\scalebox{0.9}{NAS} \\\scalebox{0.8}{(Search Space:}\\\scalebox{0.8}{Layer 1-6} \scalebox{0.8}{FFN)}}} & 1-2 (top1) & \multirow{3}{*}{\shortstack{\scalebox{0.9}{FFN}}} & \multirow{3}{*}{\shortstack{\scalebox{0.9}{\xmark}}} & \multirow{3}{*}{\shortstack{\scalebox{0.9}{\cmark}}} & \multirow{3}{*}{\shortstack{\scalebox{0.9}{\xmark}}} & {47.1} & 16.0 & 29.7 & 27.4 & {24.4} & {48.0} & 15.8 & 34.4 & 30.5 & {26.9} \\
     18 & & & 1 (top2) & & & & & {46.6} & 16.0 & 29.7$^\dag$ & 27.4 & {24.4} & {47.3} & 15.8 & 34.3$^\dag$ & 30.4 & {26.8} \\
     19 & & & 2 (top3) & & & & & {46.9} & 16.0 & 29.7 & 27.4 & {24.4} & {47.8} & 15.8 & 34.4 & 30.5 & {26.9} \\
     \hline
     20 & \multirow{2}{*}{\shortstack{\scalebox{0.9}{\ \ +Transformer}}} & Manual & 1 & \scalebox{0.9}{FFN} & \multirow{1}{*}{\shortstack{\scalebox{0.9}{\xmark}}} & \multirow{1}{*}{\shortstack{\scalebox{0.9}{\cmark}}} & \multirow{1}{*}{\shortstack{\scalebox{0.9}{\xmark}}} & \revision{46.6} & \bf15.9$^\dag$ & {\bf29.6}$^\dag$ & \bf27.3$^\dag$ & {\bf24.3}$^\dag$ & \revision{47.2} & \bf15.7$^\dag$ & \bf34.3$^\dag$ & \bf30.4$^\dag$ & {\bf26.8}$^\dag$ \\
     \cline{3-18}
     21 & & \multirow{1}{*}{\shortstack{NAS}} & 1-2 & \scalebox{0.9}{FFN} & \multirow{1}{*}{\shortstack{\scalebox{0.9}{\xmark}}} & \multirow{1}{*}{\shortstack{\scalebox{0.9}{\cmark}}} & \multirow{1}{*}{\shortstack{\scalebox{0.9}{\xmark}}} & \revision{46.9} & \bf15.9$^\dag$ & 29.7$^\dag$ & \bf34.3$^\dag$ & {\bf24.3}$^\dag$ & \revision{47.8} & \bf15.7$^\dag$ & \bf27.3$^\dag$ & \bf30.4 & {\bf26.8}$^\dag$ \\
     \hline
     \hline
     22 & \multirow{10}{*}{\shortstack{\scalebox{0.9}{GP Transformer}\\\scalebox{0.9}{+3g}}} & \multirow{6}{*}{Manual} & 1 & \multirow{6}{*}{\scalebox{0.9}{FFN}} & \multirow{6}{*}{\shortstack{\scalebox{0.9}{\cmark}}} & \multirow{6}{*}{\shortstack{\scalebox{0.9}{\cmark}}} & \multirow{6}{*}{\shortstack{\scalebox{0.9}{\xmark}}} & {47.0} & 15.9$^\dag$ & 29.7$^\dag$ & 27.4 & {24.3} & {48.1} & 15.8 & 34.4 & 30.4 & {26.9} \\
     23 & & & 1-2 & & & & & {47.4} & 16.0 & 29.8 & 27.4 & {24.4} & {48.5} & 15.9 & 34.4 & 30.5 & {26.9} \\
     24 & & & 1-3 & & & & & {47.6} & 16.0 & 29.8 & 27.5 & {24.4} & {48.8} & 15.9 & 34.5 & 30.5 & {27.0} \\
     25 & & & 1-4 & & & & & {47.8} & 16.1 & 29.8 & 27.5 & {24.5} & {49.0} & 15.9 & 34.4 & 30.5 & {26.9} \\
     26 & & & 1-5 & & & & & {48.3} & 16.2 & 29.9 & 27.5 & {24.5} & {49.4} & 15.9 & 34.5 & 30.6 & {27.0} \\
     27 & & & 1-6 & & & & & {48.1} & 16.1 & 29.9 & 27.6 & {24.5} & {49.5} & 16.0 & 34.7 & 30.6 & {27.1} \\
     \cline{3-18}
     28 & & \multirow{3}{*}{\shortstack{\scalebox{0.9}{NAS}\\\scalebox{0.8}{(Search Space:}\\\scalebox{0.8}{Layer 1-6} \scalebox{0.8}{FFN)}}} & 1,2,6 (top1) & \multirow{3}{*}{\scalebox{0.9}{FFN}} & \multirow{3}{*}{\shortstack{\scalebox{0.9}{\cmark}}} & \multirow{3}{*}{\shortstack{\scalebox{0.9}{\cmark}}} & \multirow{3}{*}{\shortstack{\scalebox{0.9}{\xmark}}} & {46.3} & 16.0 & 29.7$^\dag$ & 27.4 & {24.4} & {47.1} & 15.8 & 34.4 & 30.4 & {26.9} \\
     29 & & & 1,6 (top2) & & & & & {46.5} & 16.0 & 29.8 & 27.4 & {24.4} & {47.2} & 15.8 & 34.4 & 30.5 & {26.9} \\
     30 & & & 1,2,5,6 (top3) & & & & & {46.4} & 16.0 & 29.8 & 27.5 & {24.4} & {47.1} & 15.8 & 34.4 & 30.5 & {26.9} \\
     \hline
     31 &\multirow{2}{*}{\shortstack{\scalebox{0.9}{\ \ +Transformer}}} & Manual & 1 & \scalebox{0.9}{FFN} & \multirow{1}{*}{\shortstack{\scalebox{0.9}{\cmark}}} & \multirow{1}{*}{\shortstack{\scalebox{0.9}{\cmark}}} & \multirow{1}{*}{\shortstack{\scalebox{0.9}{\xmark}}} & \revision{46.4} & \bf15.9$^\dag$ & {\bf29.6}$^\dag$ & \bf27.3$^\dag$ & {\bf24.3}$^\dag$ & \revision{47.2} & \bf15.7$^\dag$ & \bf34.3$^\dag$ & \bf30.3$^\dag$ & {\bf26.8}$^\dag$ \\
     \cline{3-18}
     32 & & \multirow{1}{*}{\shortstack{NAS}} & 1,2,6 & \scalebox{0.9}{FFN} & \multirow{1}{*}{\shortstack{\scalebox{0.9}{\cmark}}} & \multirow{1}{*}{\shortstack{\scalebox{0.9}{\cmark}}} & \multirow{1}{*}{\shortstack{\scalebox{0.9}{\xmark}}} & \revision{46.6} & \bf15.9$^\dag$ & 29.7$^\dag$ & \bf27.3$^\dag$ & {24.3}$^\dag$ & \revision{47.2} & \bf15.7$^\dag$ & \bf34.3$^\dag$ & \bf30.3$^\dag$ & {\bf26.8}$^\dag$ \\
    \hline
    \hline
     33 & \multirow{3}{*}{\shortstack{\scalebox{0.9}{Variational}\\\scalebox{0.9}{ Transformer}\\\scalebox{0.9}{+3g}}} & \multirow{3}{*}{Manual} & 1 & \multirow{3}{*}{\shortstack{\scalebox{0.9}{Hidden}\\\scalebox{0.9}{output}}} & \multirow{3}{*}{\shortstack{\scalebox{0.9}{\xmark}}} & \multirow{3}{*}{\shortstack{\scalebox{0.9}{\xmark}}} & \multirow{3}{*}{\shortstack{\scalebox{0.9}{\cmark}}} & {48.3} & 16.1 & 29.8 & 27.5 & {24.5} & {49.5} & 15.8 & 34.5 & 30.5 & {26.9} \\
     34 & & & 2 & & & & & {48.5} & 16.1 & 29.8 & 27.6 & {24.5} & {49.6} & 15.8 & 34.6 & 30.6 & {26.9} \\
     35 & & & 1,2 & & & & & {48.4} & 16.1 & 29.8 & 27.5 & {24.5} & {49.4} & 15.8 & 34.5 & 30.6 & {27.0} \\
     \hline
     36 & \multirow{1}{*}{\shortstack{\scalebox{0.9}{\ \ +Transformer}}} & \multirow{1}{*}{Manual} & 1 & \multirow{1}{*}{\shortstack{\scalebox{0.8}{Hidden output}}} & \multirow{1}{*}{\shortstack{\scalebox{0.9}{\xmark}}} & \multirow{1}{*}{\shortstack{\scalebox{0.9}{\xmark}}} & \multirow{1}{*}{\shortstack{\scalebox{0.9}{\cmark}}} & \revision{47.3} & 16.0 & 29.7 & 27.4 & {24.4} & \revision{48.4} & \bf15.7$^\dag$ & 34.4 & 30.5 & {26.9} \\
    \bottomrule
    \end{tabular}}
    \vspace{-0.3em}
    \label{table:4}
\end{table*}

\begin{table*}[htb]
    \centering
    \scriptsize
    \captionsetup{font={small,it}}
    \caption{{Perplexity and WER(\%) of the baseline {\bf LSTM-RNN+Transformer} interpolated LM and the 5-way interpolated model between the baseline 3-gram, LSTM-RNN, Transformer LMs and their respective best Bayesian, GP and variational counterparts on AMI {\bf dev} and {\bf eval} sets of {\bf ihm}, {\bf sdm1} and {\bf mdm8} conditions.
    "$\dag$" denotes statistically significant results were obtained over the {\bf LSTM-RNN+Transformer+3g} interpolation (\revision{line 2}).}}
    \vspace{-0.5em}
    \resizebox{.85\textwidth}{!}{
    \begin{tabular}{c|l|c|cccc|c|cccc}
    \toprule
     \multirow{2}{*}{\bf ID} & \multirow{2}{*}{\bf LMs} & \bf PPL & \multicolumn{4}{c|}{\revisions{{\bf WER(\%)} of {\bf dev}}} & \bf PPL & \multicolumn{4}{c}{\revisions{{\bf WER(\%)} of {\bf eval}}} \\
     & & \multicolumn{1}{c|}{{\bf (dev)}} & \bf ihm & \bf sdm1 & \bf mdm8 & \bf Avg. & \multicolumn{1}{c|}{{\bf (eval)}} & \bf ihm & \bf sdm1 & \bf mdm8 & \bf Avg. \\
     \hline
     \hline
     1 & \revision{Oracle} & - & \revision{10.3} & \revision{20.9} & \revision{19.1} & \revision{16.8} & - & \revision{9.9} & \revision{25.1} & \revision{21.8} & \revision{18.9} \\
     \hline
     2 & {Transformer +LSTM +3g (line 2, Table \ref{table:3} + line 2, Table \ref{table:4})} & \revision{54.8} & {15.9} & {29.5} & {27.3} & {24.2} & \revision{55.1} & {15.6} & {34.3} & {30.4} & {26.8} \\
     \hline
     3 & {\revision{All NNLMs(L1) (line 4, Table \ref{table:3} + line 4, Table \ref{table:4})}} & \revision{50.2} & \revision{15.8} & \revision{29.4} & \revision{27.3} & \revision{24.2} & \revision{52.0} & \revision{15.6} & \revision{34.2} & \revision{30.3} & \revision{26.7} \\
     4 & {\revision{All NNLMs(L2) (line 6, Table \ref{table:3} + line 6, Table \ref{table:4})}} & \revision{50.1} & \revision{15.8} & \revision{29.4} & \revision{27.1} & \revision{24.1} & \revision{51.4} & \revision{15.5} & \revision{34.2} & \revision{30.2} & \revision{26.6} \\
     5 & {\revision{All NNLMs(MAP) (line 8, Table \ref{table:3} + line 8, Table \ref{table:4})}} & \revision{48.5} & \revision{15.8} & \revision{29.4} & \revision{27.1} & \revision{24.0} & \revision{48.8} & \revision{15.4} & \revision{34.1} & \revision{30.2} & \revision{26.6} \\
    \hline
    \hline
     6 & \multirow{1}{*}{\shortstack{{All Bayesian NNLMs (\revision{line 24, Table \ref{table:3} + line 21, Table \ref{table:4}}})}} & \revision{46.4} & {\bf 15.6}$^\dag$ & {\bf {29.2}}$^\dag$ & {\bf 26.9}$^\dag$ & {\bf 23.9}$^\dag$ & \revision{47.9} & {\bf 15.3}$^\dag$ & {\bf {33.8}}$^\dag$ & {\bf {29.9}}$^\dag$ & {\bf 26.3}$^\dag$ \\
     7 & \multirow{1}{*}{\shortstack{{All GP NNLMs (\revision{line 44, Table \ref{table:3} + line 32, Table \ref{table:4}})}}} & \revision{46.5} & {15.7}$^\dag$ & {{29.3}} & {27.1}$^\dag$ & {24.0}$^\dag$ & \revision{48.2} & {\bf 15.3}$^\dag$ & {{34.0}}$^\dag$ & {{30.1}}$^\dag$ & {26.5}$^\dag$ \\
     8 & \multirow{1}{*}{\shortstack{{All Variational NNLMs (\revision{line 48, Table \ref{table:3} + line 36, Table \ref{table:4}})}}} & \revision{47.4} & {15.8} & {{29.4}} & {27.1}$^\dag$ & {24.1} & \revision{48.4} & {15.4}$^\dag$ & {{34.2}} & {{30.2}}$^\dag$ & {26.6}$^\dag$ \\
     \hline
    \bottomrule
    \end{tabular}}
    \vspace{-1.9em}
    \label{table:5}
\end{table*}

\section{{Experiments}}
\label{sec:exp}
In this section the performance of various Bayesian learning based LSTM-RNN and Transformer LMs are evaluated on the speech recognition system using state-of-the-art LF-MMI sequence trained time delay neural networks (TDNNs) \cite{povey2016purely} featuring speed perturbation based data augmentation and i-Vector speaker adaptation \cite{saon2013speaker}. 
Audio-visual multi-channel beamforming and recognition \cite{yu2021audio} was also used when processing overlapping speech. 
In Sec.\ref{sec:exp_ami}, the first set of experiments conducted on the AMI meeting room data \cite{hain2006ami} were presented. 
The experiments were designed to provide a detailed side by side analysis over the three Bayesian learning based NNLM approaches presented in Sec.\ref{sec:BLN}, while following the implementation issues and the best configurations of Bayesian NNLMs discussed in Sec.\ref{sec:Imp}. 
A second set of experiments are conducted in Sec.\ref{sec:exp_lrs2} on an audio-visual multi-channel overlapped speech recognition task based on Oxford-BBC Lip Reading Sentences 2 (LRS2) corpus \cite{chung2017lip} to further evaluate the performance of Bayesian NNLMs.

All the LSTM-RNN LMs investigated in this paper consist of 2 LSTM layers while both the input word embedding and hidden layer sizes were set as 1024. 
All the Transformer LMs contain 6 Transformer layers. 
The dimensionality of all query, key and value embedding layers were set as 512, and the hidden vector dimensionality was set as 4096. 
Both LSTM-RNN and Transformer LMs were implemented using PyTorch \cite{paszke2017automatic}. 
All NNLMs were trained using a single NVIDIA Tesla V100 Volta GPU card. 
SGD parameter update in a mini-batch mode (32 sentences per batch) was used. 
Dropout regularization was also used and the dropout rate was set at 0.2 in all experiments. 
\revision{A set of baseline NNLMs using L1 \cite{tibshirani1996regression} or L2 \cite{tikhonov1963solution} regularization, or maximum a posteriori (MAP) estimation \cite{chien2014bayesian,chien2015bayesian} with the same priors used by various forms of Bayesian NNLMs in this work, were also presented as the contrasts to those Bayesian estimated NNLMs.
}
{Initial learning rate settings of 5 and 0.1 were used for the baseline LSTM-RNN and Transformer LMs respectively. 
A smaller initial learning rate of 0.1 and 0.01 were used for the Bayesian estimated LSTM-RNN and Transformer LMs. 
For all models, the learning rate was halved during SGD update whenever the perplexity reduction was not obtained on the validation set.
These baseline NNLM settings are based on those provided by the Kaldi recipe\footnote{\href{https://github.com/kaldi-asr/kaldi/blob/master/egs/swbd/s5c/local/pytorchnn/run\_nnlm.sh}{Kaldi: egs/swbd/s5c/local/pytorchnn/run\_nnlm.sh}} before being further fine-tuned.
}

\revision{During performance evaluation, by default all the NNLMs, baseline or Bayesian estimated, were all linearly interpolated with the $n$-gram back-off LM. 
For all the Bayesian estimated NNLMs of this paper, there are optionally further 3-way interpolated with the $n$-gram and their respective baseline NNLMs using standard point estimate.
For example, for the Bayesian LSTM-RNN LM produced word probability, $P_{\rm {blstm}}(\cdot)$, the 3-way interpolated LM probability with the baseline $n$-gram $P_{\rm {ng}}(\cdot)$ and point estimated LSTM-RNN $P_{\rm {lstm}}(\cdot)$ LMs is given by}
\vspace{-0.3em}
\begin{align}
\label{eq:lm_inter}
    \color{black}{\small P\left (\boldsymbol w_t|\boldsymbol w_{1}^{t-1}\right )=}&\color{black}{\small \lambda_1 P_{\rm {ng}} \left (\boldsymbol w_t|\boldsymbol w_{1}^{t-1}\right )+ \lambda_2 P_{\rm {lstm}}\left (\boldsymbol w_t|\boldsymbol w_{1}^{t-1}\right )}\nonumber\\
    &\color{black}{\small +(1-\lambda_1-\lambda_2) P_{\rm {blstm}}\left (\boldsymbol w_t|\boldsymbol w_{1}^{t-1}\right )}
\end{align}
\revision{where $\lambda_1, \lambda_2$ denote the interpolation weights that are optimized by minimizing the perplexity on the held-out development data set using the expectation maximization (EM) algorithm.
The above LM interpolation also allows a more powerful multi-way combination between $n$-gram, LSTM-RNN and Transformer LMs using point estimate or Bayesian learning to be performed.}
Statistical significance test was conducted at level $\alpha$$=$$0.05$ based on matched pairs sentence segment word error (MAPSSWE) \cite{gillick1989some} for recognition performance analysis.
\vspace{-0.5em}








\vspace{-0.1em}

\subsection{{Experiments on AMI Meeting Room Data}}
\label{sec:exp_ami}
\noindent
{\bf {System Description:}}
The Augmented Multi-party Interaction (AMI) speech corpus consists of approximately 100 hours of audio data collected using both headset microphone and distant microphone arrays from the meeting environment. 
Following the Kaldi recipe\footnote{\href{https://github.com/kaldi-asr/kaldi/blob/master/egs/swbd/s5c/local/chain/tuning/run\_tdnn\_7q.sh}{Kaldi: egs/swbd/s5c/local/chain/tuning/run\_tdnn\_7q.sh}}, three LF-MMI trained \cite{povey2016purely} acoustic models with speech perturbation based data augmentation and i-Vector based speaker adaptation \cite{saon2013speaker} were then constructed. 
The AMI 8.9-hour {\bf dev} and 8.7-hour {\bf eval} sets recorded under close talking microphone ({\bf ihm}), single distant microphone ({\bf sdm}) and multiple distant microphones ({\bf mdm}) were used. 
A 47K word recognition lexicon was used. 
Various LSTM-RNN and Transformer LMs were trained on a combined data set of 15M words including both the AMI and Fisher transcripts, before being used to rescore the 3-gram LM produced N-best lists (N = 20) for WER performance evaluation. 

\begin{figure*}[t]
\newcommand{\tabincell}[2]{\begin{tabular}{@{}#1@{}}#2\end{tabular}}
\setlength{\parskip}{0.0ex}

  \centering
  \vspace{-0.5em}
  \centerline{\includegraphics[width=17.8cm]{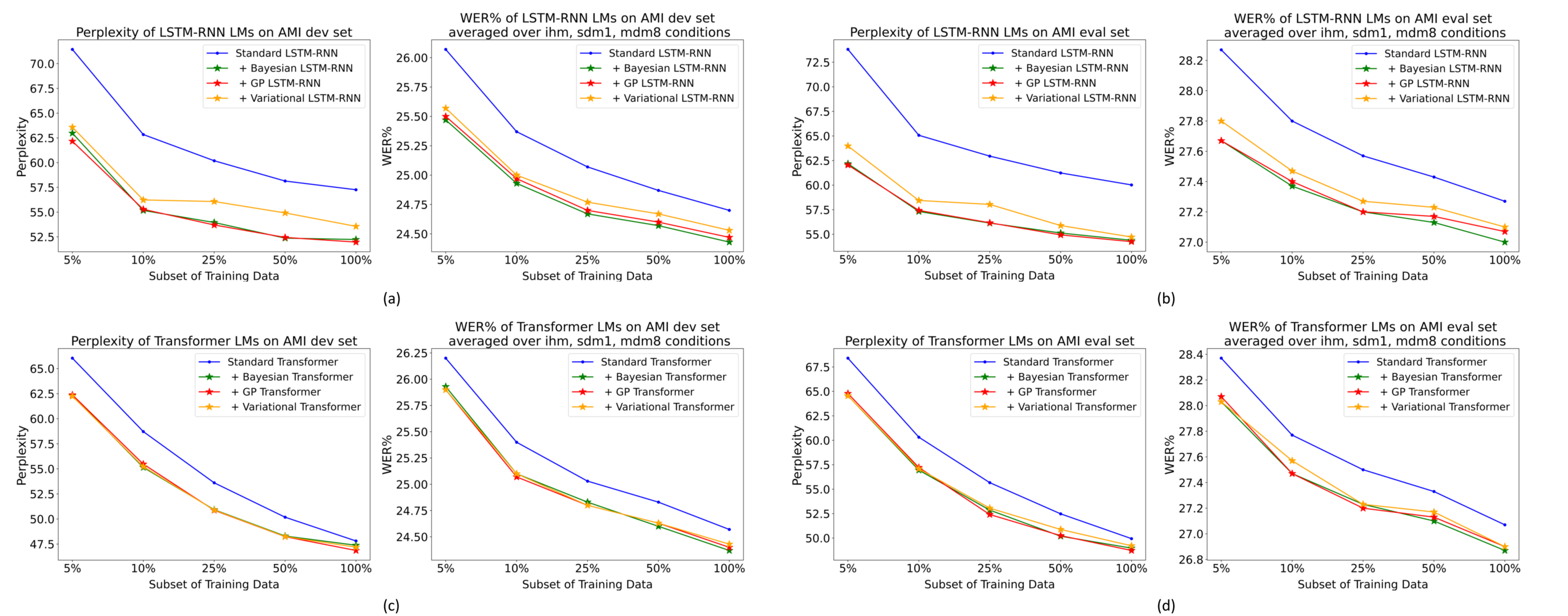}}

\vspace{-0.7em}
\captionsetup{font={small,it}}
\caption{{Performance contrasts between baseline, Bayesian, GP and Variational counterparts for both {\bf LSTM-RNN} and {\bf Transformer} LMs trained on randomly selected subsets of training data (varying from 5\% to 100\% of the 15M word AMI+SWBD training set): 
perplexity and average WER\% measured across {\bf ihm}, {\bf sdm} and {\bf mdm} conditions on the AMI {\bf dev} (a) and {\bf eval} (b) data for {\bf LSTM-RNN} LMs; perplexity and average WER\% similarly measured on the AMI {\bf dev} (c) and {\bf eval} (d) data for {\bf Transformer} LMs. 
}}
\vspace{-1.0em}
\label{fig:ami_data_size}
\end{figure*}

\begin{figure*}[t]
\newcommand{\tabincell}[2]{\begin{tabular}{@{}#1@{}}#2\end{tabular}}
\setlength{\parskip}{0.0ex}

  \centering
  \centerline{\includegraphics[width=17.8cm]{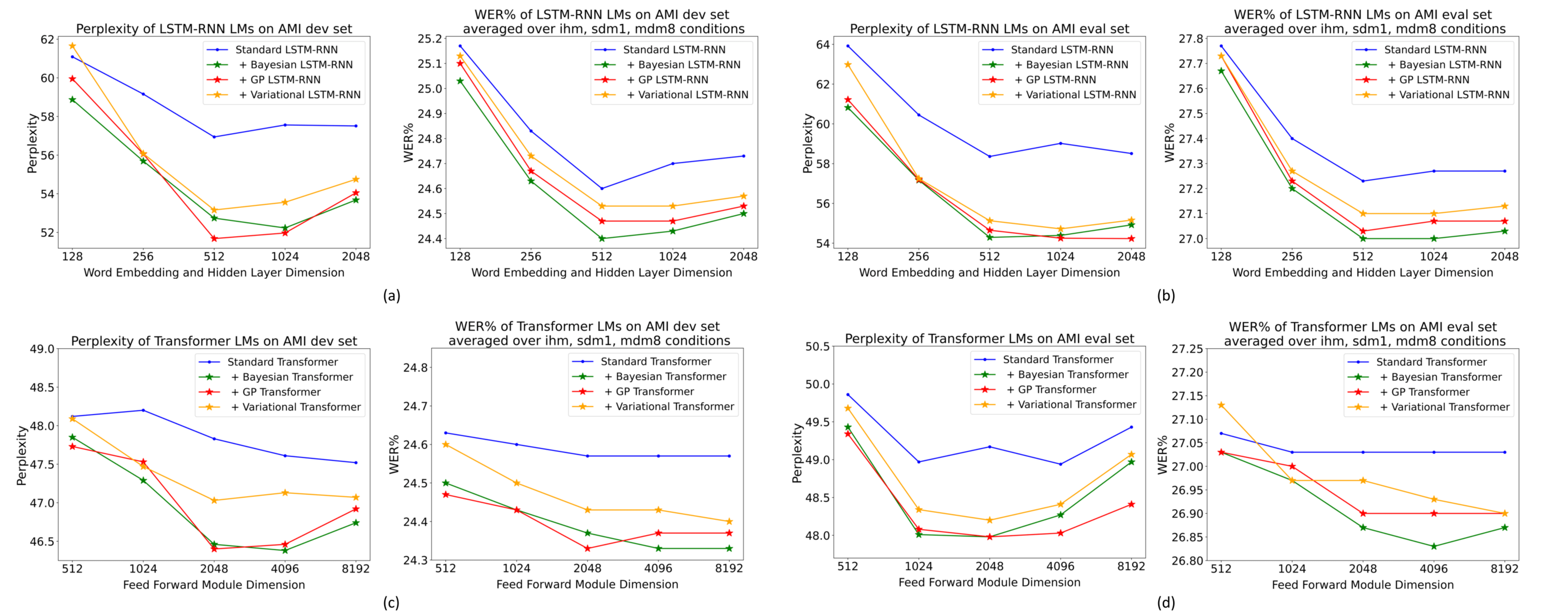}}

\vspace{-0.7em}
\captionsetup{font={small,it}}
\caption{{Performance contrasts between baseline, Bayesian, GP and Variational counterparts for both {\bf LSTM-RNN} and {\bf Transformer} LMs trained in different model sizes (varying from 128 to 2048 on both word embedding and hidden layer dimensionality for {\bf LSTM-RNNs} and from 512 to 4096 on feed forward module dimensionality for {\bf Transformers}): 
perplexity and average WER\% measured across {\bf ihm}, {\bf sdm} and {\bf mdm} conditions on the AMI {\bf dev} (a) and {\bf eval} (b) data for {\bf LSTM-RNN} LMs; perplexity and average WER\% similarly measured on the AMI {\bf dev} (c) and {\bf eval} (d) data for {\bf Transformer} LMs. 
}}
\vspace{-1.8em}
\label{fig:ami_model_size}
\end{figure*}



\noindent
\vspace{0.1em}
{\bf {Experiments on LSTM-RNN LMs:}}
Table \ref{table:3} presents the perplexity and WER performance of various Bayesian, GP and variational LSTM-RNN LMs. 
Several trends can be found:

\vspace{-0.1em}
1) Irrespective of the precise form of Bayesian modeling being used, Bayesian (\revision{line} \revision{9 to 24}), Gaussian Process (\revision{line} \revision{25 to 44}) or variational (\revision{line} \revision{45 to 48}) approaches, a general trend can be observed that Bayesian estimation produced consistent perplexity and WER reductions over the baseline LSTM-RNN LM (line 2) using \revisions{point estimated} parameters. 
In particular, the majority of the Bayesian and GP LSTM-RNN LMs (\revision{line} \revision{9 to 44}) with Bayesian or GP modeling applied to various network internal locations that were either manually or automatically selected using NAS, produced statistically significant ($\alpha$$=$$0.05$) WER reductions over the baseline LSTM-RNN LM (line 2). {Performance improvements were also obtained over the comparable L1 (line 3, 4), L2 (line 5, 6) regularized or MAP estimated (line 7, 8) LSTM-RNN LMs.}

2) A more detailed ablation study on the effect of applying Bayesian uncertainty modeling to different LSTM unit internal components (previously shown in Fig.\ref{fig:lstm}) on performance is shown for Bayesian (\revision{line} \revision{9 to 22}) and GP (\revision{line} \revision{25 to 42}) LSTM-RNN LMs respectively.
The results in Table \ref{table:3} suggest that applying Bayesian estimation to only the LSTM cell activation parameters of both hidden layers (\revision{line} \revision{20 and 24}), auto-configured as the 1-best location by the NAS approach of Sec.\ref{sec:uncertainty}, produced the best performance in comparison against other locations. 
Similarly applying GP modeling only to 
the hidden vector input gate ("h-Gate" in left bottom corner, in purple of Fig.\ref{fig:lstm} (a)) (\revision{line} \revision{40 and 44}) of both hidden layers produced the best performance for GP LSTM-RNN LMs.

\vspace{-0.1em}
3) Among three Bayesian modeling approaches, the performance of Bayesian and GP \revision{LSTM-RNN} LMs (\revision{line} \revision{23, 24} \revision{and} \revision{43, 44}) consistently outperform that of the comparable variational LSTM-RNN LMs (\revision{line} \revision{48}) which consider the uncertainty associated with the hidden output vectors. 
Between Bayesian and GP LSTM-RNN LMs, the results in Table \ref{table:3} also suggest that modeling the additional uncertainty on this data over the activation function choices inside the classic, expert crafted neural structure of LSTM units using GP {LSTM-RNN} LMs (\revision{line} \revision{43, 44}) brings no further improvements over Bayesian LSTM-RNNs (\revision{line} \revision{23, 24}) that consider the uncertainty over model parameters only.

4) The best performance was obtained using a 3-way interpolation between the 3-gram, \revisions{point estimate} parameter LSTM-RNN and Bayesian LSTM-RNN LMs (\revision{line} \revision{24}) with the precise position of applying Bayesian estimation auto-configured using NAS. 
Perplexity reduction of 13 points and WER reductions of 0.5\%-0.7\% absolute (1.7\%-3.7\% relative) were obtained over the baseline LSTM-RNN LM (line 2).

\noindent
{\bf {Experiments on Transformer LMs:}}
A set of experiments that are comparable to those previously shown in Table \ref{table:3} for LSTM-RNN LMs were then conducted for Transformer LMs. 
These are shown in Table \ref{table:4} with the following trends:

1) After interpolating the Bayesian (\revision{line} \revision{20, 21}), Gaussian Process (\revision{line} \revision{31, 32}) or variational (\revision{line} \revision{36}) Transformer LM with the 3-gram and baseline Transformer LM, a general trend is that Bayesian estimation produced small, but consistent and statistically significant ($\alpha$$=$$0.05$) WER reductions of 0.2\% to 0.3\% absolute over the baseline Transformer LM (line 2) using \revisions{point estimated} parameters across all  channel conditions. \revision{Performance improvements were also obtained over the comparable L1 (line 3, 4), or L2 (line 5, 6) regularized or MAP estimated (line 7, 8) Transformer LMs.}

2) The above improvements from Bayesian learning are notably smaller than those previously found on the comparable LSTM-RNN LM experiments in Table \ref{table:3}. This may be attributed to the larger modeling uncertainty that the LSTM-RNN LMs exhibit on this task when compared with the Transformer LM. 

3) The best Bayesian Transformer LM performance was obtained using a 3-way interpolation between the baseline 3-gram, \revisions{point estimate} parameter Transformer and GP Transformer LMs (\revision{line} \revision{31, 32}) with the precise location of applying GP estimation either manually set as the lowest positioned feedforward layer, or auto-configured using NAS.

\begin{table*}[!tp]
    \centering
    \scriptsize
    \vspace{-0.2em}
    \captionsetup{font={small,it}}
    \caption{Perplexity and WER\% of the baseline 4-gram (4g), {\bf LSTM-RNN} and {\bf Transformer} LMs with \revision{standard point estimate}, \revision{and using L1, or L2 regularization or MAP estimation}, and their Bayesian, GP and variational counterparts with various forms of uncertainty modeling on the {LRS2} {\bf Test} set of “clean”, and overlapping conditions obtained using “TF masking”, “Filter \& Sum” or “MVDR” based audio-visual neural beamforming. 
    "+4g", "+LSTM" and "+ Transformer" denote the results of further interpolation with the baseline 4-gram, {\bf LSTM-RNN} and {\bf Transformer} LMs. 
    "$\dag$", "$\ddagger$" and "$\star$" denote statistically significant WER reductions were obtained over the baseline {\bf LSTM-RNN} (\revision{line 3}), {\bf Transformer} (\revision{line 4}) LMs and {\bf LSTM-RNN}+{\bf Transformer} interpolation (\revision{line 5}) respectively.
    }
    \vspace{-0.5em}
    \resizebox{.85\textwidth}{!}{
     \begin{tabular}{c|l|ccc|c|c|c|c|c|c}
    \toprule
     \multirow{2}{*}{\bf ID} & \multirow{2}{*}{\bf LM} & \multicolumn{3}{c|}{\bf Bayesian} & \bf PPL & \multicolumn{5}{c}{\revisions{{\bf WER(\%)}}} \\
     & & \bf Design & \bf Layer & \bf Position & \revisions{\bf (Test)} & {\bf clean} & {\bf TF masking} & {\bf Filter \& Sum} & {\bf MVDR} & {\bf Avg.} \\
     \hline
     \hline
     1 & \revision{Oracle} & \multicolumn{3}{c|}{\multirow{20}{*}{\shortstack{Not Applied}}} & {-} & \revision{1.7} & \revision{5.8} & \revision{5.4} & \revision{5.2} & \revision{4.5} \\
     \cline{1-2}\cline{6-11}
     2 & 4-gram & \multicolumn{3}{c|}{} & {77.0} & {8.5} & {17.7} & {16.6} & {16.1} & {14.7} \\
     3 & LSTM +4g & & & & {65.7} & {5.8} & {13.3} & {12.1} & {11.8} & {10.7} \\
     4 & Transformer +4g & & & & {66.3} & {5.7} & {12.7} & {12.1} & {11.9} & {10.6} \\
     5 & Transformer +LSTM +4g & & & & \revision{65.8} & {5.3} & {12.2} & {11.3} & {11.6} & {10.1} \\
     \cline{1-2}\cline{6-11}
     6 & \revision{LSTM(L1) +4g} & & & & \revision{65.5} & \revision{5.7} & \revision{13.2} & \revision{12.1} & \revision{11.8} & \revision{10.7} \\
     7 & \ \ \ \ \revision{+LSTM} & & & & \revision{65.2} & \revision{5.5} & \revision{12.9} & \revision{11.8} & \revision{11.7} & \revision{10.5} \\
     8 & \revision{Transformer(L1) +4g} & & & & \revision{66.1} & \revision{5.7} & \revision{12.6} & \revision{12.0} & \revision{11.9} & \revision{10.6} \\
     9 & \ \ \ \ \revision{+Transformer} & & & & \revision{65.7} & \revision{5.5} & \revision{12.4} & \revision{11.7} & \revision{11.8} & \revision{10.3} \\
     \cline{2-2}\cline{6-11}
     10 & \revision{LSTM(L2) +4g} & & & & \revision{65.4} & \revision{5.7} & \revision{13.1} & \revision{12.0} & \revision{11.8} & \revision{10.7} \\
     11 & \ \ \ \ \revision{+LSTM} & & & & \revision{65.0} & \revision{5.5} & \revision{12.8} & \revision{11.7} & \revision{11.6} & \revision{10.4} \\
     12 & \revision{Transformer(L2) +4g} & & & & \revision{65.9} & \revision{5.6} & \revision{12.6} & \revision{11.8} & \revision{11.9} & \revision{10.5} \\
     13 & \ \ \ \ \revision{+Transformer} & & & & \revision{65.2} & \revision{5.4} & \revision{12.4} & \revision{11.6} & \revision{11.7} & \revision{10.3} \\
     \cline{2-2}\cline{6-11}
     14 & \revision{LSTM(MAP) +4g} & & & & \revision{65.3} & \revision{5.6} & \revision{13.1} & \revision{12.0} & \revision{11.8} & \revision{10.6} \\
     15 & \ \ \ \ \revision{+LSTM} & & & & \revision{64.9} & \revision{5.5} & \revision{12.8} & \revision{11.7} & \revision{11.6} & \revision{10.4} \\
     16 & \revision{Transformer(MAP) +4g} & & & & \revision{65.8} & \revision{5.6} & \revision{12.5} & \revision{11.8} & \revision{11.9} & \revision{10.5} \\
     17 & \ \ \ \ \revision{+Transformer} & & & & \revision{65.0} & \revision{5.5} & \revision{12.4} & \revision{11.5} & \revision{11.7} & \revision{10.3} \\
     \cline{2-2}\cline{6-11}
     18 & \revision{All NNLMs(L1) (line 7 + 9)} & & & & \revision{64.9} & \revision{5.3} & \revision{12.2} & \revision{11.2} & \revision{11.4} & \revision{10.0} \\
     19 & \revision{All NNLMs(L2) (line 11 + 13)} & & & & \revision{64.8} & \revision{5.2} & \revision{12.2} & \revision{11.1} & \revision{11.4} & \revision{10.0} \\
     20 & \revision{All NNLMs(MAP) (line 15 + 17)} & & & & \revision{64.6} & \revision{5.2} & \revision{12.1} & \revision{11.1} & \revision{11.3} & \revision{9.9} \\
     \hline
     \hline
     21 & \multirow{2}{*}{{Bayesian LSTM +4g}} & Manual & 2 & All gates & {65.5} & {5.3}$^\dag$ & {12.4}$^\dag$ & {11.8} & {12.0} & {10.4} \\
     22 & & NAS & 1,2 & {Cell input} & {65.5} & {5.4}$^\dag$ & {12.5}$^\dag$ & {11.8} & {11.9} & {10.4}$^\dag$ \\
     \hline
     23 & \multirow{2}{*}{{GP LSTM +4g}} & Manual & 1,2 & {Cell input} & {65.6} & {5.2}$^\dag$ & {12.5}$^\dag$ & {11.7}$^\dag$ & {12.2} & {10.4}$^\dag$ \\
     24 & & NAS & 1,2 & h-Gate & {65.2} & {5.3}$^\dag$ & {{12.4}}$^\dag$ & {11.8} & {12.2} & {10.4} \\
     \hline
     25 & \multirow{1}{*}{\shortstack{{V-LSTM +4g}}} & Manual & 1,2 & Hidden output & {62.4} & {5.5} & {12.6}$^\dag$ & {11.7}$^\dag$ & {11.6} & {10.5} \\
     \hline
     \hline
     26 & \multirow{2}{*}{\shortstack{{Bayesian Transformer +4g}}} & Manual & 1 & \multirow{2}{*}{\shortstack{FFN}} & {66.7} & {5.5} & {12.5} & {11.8} & {12.3} & {10.5} \\
     27 & & NAS & 1,2 & & {65.0} & {5.2}$^\ddagger$ & {12.3}$^\ddagger$ & {11.5}$^\ddagger$ & {12.3} & {10.3}$^\ddagger$ \\
     \hline
     28 & \multirow{2}{*}{\shortstack{{GP Transformer +4g}}} & Manual & 1 & \multirow{2}{*}{\shortstack{FFN}} & {64.7} & {5.4}$^\ddagger$ & {12.4} & {11.5}$^\ddagger$ & {11.9} & {10.3} \\
     29 & & NAS & 1,2,6 & & {65.4} & {5.5} & {12.6} & {11.8} & {12.0} & {10.6} \\
     \hline
     30 & {V-Transformer +4g} & Manual & 1 & Hidden output & {66.1} & {5.6} & {12.7} & {11.9} & {12.4} & {10.7} \\
     \hline
     \hline
     31 & \multirow{2}{*}{\shortstack{{Bayesian LSTM +4g +LSTM}}} & Manual & 2 & All gates & \revision{64.4} & {5.1}$^\dag$ & {12.1}$^\dag$ & {11.4}$^\dag$ & {\bf11.3}$^\dag$ & {\bf10.0}$^\dag$ \\
     32 & & NAS & 1,2 & {Cell input} & \revision{64.4} & {5.1}$^\dag$ & {12.1}$^\dag$ & {11.4}$^\dag$ & {\bf11.3}$^\dag$ & {\bf10.0}$^\dag$ \\
     \hline
     33 & \multirow{2}{*}{\shortstack{{GP LSTM +4g +LSTM}}} & Manual & 1,2 & {Cell input} & \revision{64.5} & {\bf5.0}$^\dag$ & {12.0}$^\dag$ & {\bf11.3}$^\dag$ & {11.6} & {10.0}$^\dag$ \\
     34 & & NAS & 1,2 & h-Gate & \revision{63.9} & {5.0}$^\dag$ & {\bf{12.0}}$^\dag$ & \revision{11.5}$^\dag$ & {11.5} & {10.0}$^\dag$ \\
     \hline
     35 & \multirow{1}{*}{\shortstack{V-LSTM +4g +LSTM}} & Manual & 1,2 & Hidden output & \revision{62.3} & {5.3}$^\dag$ & {12.0}$^\dag$ & {11.3}$^\dag$ & {11.6} & {10.1}$^\dag$ \\
     \hline
     \hline
     36 & \multirow{2}{*}{\shortstack{{Bayesian Transformer +4g +Transformer}}} & Manual & 1 & \multirow{2}{*}{\shortstack{FFN}} & \revision{65.9} & {5.2}$^\ddagger$ & {12.0}$^\ddagger$ & \revision{11.3}$^\ddagger$ & \revision{11.8} & {10.1}$^\ddagger$ \\
     37 & & NAS & 1,2 & & \revision{64.0} & {\bf5.0}$^\ddagger$ & {\bf11.8}$^\ddagger$ & {11.2}$^\ddagger$ & {11.6}$^\ddagger$ & {9.9}$^\ddagger$ \\
     \hline
     38 & \multirow{2}{*}{\shortstack{{GP Transformer +4g +Transformer}}} & Manual & 1 & \multirow{2}{*}{\shortstack{FFN}} & \revision{63.8} & {5.0}$^\ddagger$ & {11.9}$^\ddagger$ & {\bf11.1}$^\ddagger$ & {\bf11.4}$^\ddagger$ & {\bf9.9}$^\ddagger$ \\
     39 & & NAS & 1,2,6 & & \revision{64.7} & {5.2}$^\ddagger$ & \revision{12.0}$^\ddagger$ & {11.4}$^\ddagger$ & {11.6}$^\ddagger$ & {10.1}$^\ddagger$ \\
     \hline
     40 & V-Transformer +4g +Transformer & Manual & 1 & Hidden output & \revision{64.7} & {5.2}$^\ddagger$ & \revision{12.2}$^\ddagger$ & {11.4}$^\ddagger$ & {11.6}$^\ddagger$ & {10.1}$^\ddagger$ \\
    \hline
    \hline
     41 & \multirow{1}{*}{\shortstack{All Bayesian NNLMs (\revision{line 32 + 37})}} & - & - & - & \revision{64.0} & {4.8}$^\star$ & {11.7}$^\star$ & {10.8}$^\star$ & {11.0}$^\star$ & {9.6}$^\star$ \\
     42 & \multirow{1}{*}{\shortstack{All GP NNLMs (\revision{line 34 + 39})}} & - & - & - & \revision{63.7} & {\bf4.7}$^\star$ & {\bf11.4}$^\star$ & {\bf10.7}$^\star$ & {\bf10.8}$^\star$ & {\bf9.4}$^\star$ \\
     43 & \multirow{1}{*}{\shortstack{All Variational NNLMs (\revision{line 35 + 40})}} & - & - & - & \revision{64.1} & {4.9}$^\star$ & {11.8}$^\star$ & {10.8}$^\star$ & {11.2}$^\star$ & {9.7}$^\star$ \\
    \bottomrule
    \end{tabular}}
    \vspace{-1.8em}
    \label{table:6}
\end{table*}

\begin{table}[tb]
    \centering
    \scriptsize
    \captionsetup{font={small,it}}
    \caption{{
    The median parameter {\bf signal-to-noise ratio (SNR)} statistics \cite{blundell2015weight,braun2019parameter} measured over Bayesian {\bf LSTM-RNN} (B-LSTM) and {\bf Transformer} (B-Transformer) LMs trained on randomly selected subsets of training data varying from 5\% to 100\% of the 15M word AMI+SWBD training set (line 1, 2), and different model sizes varying from 128 to 2048 in terms of both the word embedding and hidden layer dimensionality for LSTM-RNNs and from 512 to 8192 in terms of feed forward module dimensionality for Transformers (line 3, 4).
    }
    }
    \vspace{-0.5em}
    \resizebox{0.46\textwidth}{!}{
     \begin{tabular}{c|l|ccccc}
    \toprule
     {\bf ID} & {\shortstack{\bf LM}} & \multicolumn{5}{c}{\bf Subset of Training Data / SNR} \\
     \hline
     1 & {\scalebox{0.9}{B-LSTM}} & {5\%/{\bf 0.3}} & {10\%/{\bf 0.4}} & {25\%/{\bf 0.5}} & {50\%/{\bf 0.7}} & {100\%/{\bf 1.1}} \\
     2 & {\scalebox{0.9}{B-Transformer}} & {5\%/{\bf 0.6}} & {10\%/{\bf 0.8}} & {25\%/{\bf 1.1}} & {50\%/{\bf 2.0}} & {100\%/{\bf 3.3}} \\
     \hline
     \hline
      & {\shortstack{\bf LM}} & \multicolumn{5}{c}{\bf Hidden (FFN) Layer Dimensionality / SNR} \\
     \hline
     3 & {\scalebox{0.9}{B-LSTM}} & {128/{\bf 5.1}} & {256/{\bf 2.6}} & {512/{\bf 1.4}} & {1024/{\bf 1.1}} & {2048/{\bf 2.0}} \\
     4 & {\scalebox{0.9}{B-Transformer}} & {512/{\bf 8.2}} & {1024/{\bf 4.3}} & {2048/{\bf 3.0}} & {4096/{\bf 3.3}} & {8192/{\bf 3.7}}  \\
    \bottomrule
    \end{tabular}}
    \vspace{-2.3em}
    \label{table:7}
\end{table}


\noindent

\noindent
{{\bf {Experiments of Uncertainty Analysis:}}
The following set of ablation studies were further conducted to evaluate Bayesian NNLMs' performance and provide modeling uncertainty analyses in Fig.\ref{fig:ami_data_size}, Fig.\ref{fig:ami_model_size} and Table \ref{table:7}:}

{1) A first ablation study shown in Fig.\ref{fig:ami_data_size} is to illustrate that statistically significant WER reductions, and perplexity improvements were consistently obtained using Bayesian learned LSTM-RNN and Transformer LMs trained on randomly selected subsets of training data (varying from 5\% to 100\% of the 15M word AMI+SWBD training set) over their respective \revisions{point estimated} baseline LMs on each subset of training data.
These two sets of experiments serve to analyze the improvements of Bayesian estimation using a series of fixed model complexity versus varying data quantity operating points.}

{2) Another set of experiments were then conducted using a different set of model complexity versus data quantity operating points by using all the AMI training data, but varying either the LSTM-RNN LM word embedding and hidden layer dimensionality, or the dimensionality of the Transformer LM feed-forward modules. 
These contrasts are shown in Fig.\ref{fig:ami_model_size}.}

{3) For Bayesian estimated LSTM-RNN and Transformer LMs constructed using the above different model complexity versus data quantity trade-off points, the median signal-to-noise ratio (SNR) of the Gaussian approximated latent NNLM model parameter distributions \cite{blundell2015weight,braun2019parameter}}
\vspace{-0.3em}
\begin{align}
    {{\small p({\boldsymbol \Theta|\mathcal D})\approx q(\boldsymbol \Theta)=\mathcal N(\boldsymbol \Theta|\boldsymbol\mu,\revisions{\boldsymbol\Sigma_{\sf diag}})}}
\end{align}
{is analyzed and shown in Table \ref{table:7} as a measure of parameter uncertainty.
The $i$-th element $\Theta_i$ in parameter $\boldsymbol \Theta$ specific signal-to-noise ratio (SNR) is defined as follows.}
\vspace{-0.5em}
\begin{align}
    {{\small \mathrm{SNR}_{\Theta_i}=\revisions{\frac{|\mu_i|}{\sigma_i}}}}
\end{align}

{A general trend can be found that either decreasing the training data quantity, or increasing the NNLM model complexity, produced a lower parameter SNR, which indicate an increase in model parameter uncertainty and risk of over-fitting.}
\vspace{-0.5em}

\subsection{{Experiments on the LRS2}}
\label{sec:exp_lrs2}

\noindent
{\bf {System Description:}}
The Oxford-BBC Lip Reading Sentences 2 (LRS2) corpus is one of the largest publicly available corpora for audio-visual speech recognition, which consists of news and talk shows extracted from BBC broadcast.
Multi-channel cocktail party overlapped speech was simulated with an 85\% average overlapping ratio based on the LRS2 corpus. 
An audio-visual multi-channel overlapped speech recognition (AVSR) system~\cite{yu2020audio,yu2021audio} was then constructed, featuring tightly integrated separation front-end and recognition back-end components jointly fine-tuned using a multi-task interpolation of the scale-invariant signal to noise ratio (Si-SNR) and LF-MMI criteria. 
An AVSR system trained on non-overlapped, clean speech, as well as another three AVSR systems trained on overlapped speech using different audio-visual multi-channel beamforming approaches including TF Masking, Filter \& Sum and Mask-based MVDR, were used for evaluation. 
The LRS2 corpus is already divided into three subsets: {\bf Pre-train}, {\bf Train-val} and {\bf Test}. 
Various baseline LSTM-RNN and Transformer LMs are trained on the combined {\bf Pre-train} and {\bf Train-Val} set containing 2.5M words using a 41K word recognition lexicon. 
The resulting NNLMs were used to rescore the 4-gram LM produced N-best lists (N = 20) for WER evaluation.

\noindent
{\bf {Experimental Results:}}
A set of experiments that are comparable to those {in Sec.\ref{sec:exp_ami}} conducted on the AMI data in Table \ref{table:3} and \ref{table:4} for LSTM-RNN and Transformer LMs are shown in {Table \ref{table:6}}.
For all the Bayesian, GP and variational LSTM-RNN and Transformer LMs, their respective best network internal positions to apply Bayesian modeling were based on those previously selected either manually or automatically using NAS in Table \ref{table:3} and \ref{table:4}. 
The following trends are found:

1) For both LSTM-RNN and Transformer LMs, using the Bayesian (\revision{line} \revision{21, 22, 26, 27, 31, 32, 36, 37}), Gaussian Process (\revision{line} \revision{23, 24, 28, 29, 33, 34, 38, 39}) or variational (\revision{line} \revision{25, 30, 35, 40}) approaches, a general trend can be observed that Bayesian estimation produced consistent perplexity and WER reductions over the corresponding baseline NNLMs (\revision{line} \revision{3 and 4}) using \revisions{point estimated}, deterministic parameters.
\revision{Performance improvements were also obtained over the comparable L1 (line 6 to 9), or L2 (line 10 to 13) regularized or MAP estimated (line 14 to 17) LSTM-RNN and Transformer LMs.}

2) Among three Bayesian modeling approaches, GP LSTM-RNN and Transformer LMs (\revision{line} \revision{23, 24, 28, 29, 33, 34, 38, 39}) on average across both the clean and three overlapped speech conditions outperform their Bayesian (\revision{line} \revision{21, 22, 26, 27, 31, 32, 36, 37}) and variational counterparts (\revision{line} \revision{25, 30, 35, 40}) considering the uncertainty over either model parameters or hidden output vectors only. 
This is expected as on this LRS2 task the training data size is reduced to 2.5M words, 16.7\% of the 15M word AMI training data, while the LSTM-RNN or Transformer LM model complexity {remains} the same. 
Hence, a larger risk of over-fitting arises and requires more powerful uncertainty modeling over both the hidden activation functions and their weight parameters using GP NNLMs.

\revisions{3) No statistically significant ($\alpha$$=$$0.05$) WER difference was obtained among the best performing Bayesian/GP LSTM-RNN LMs (line 21, 22 and 24) and variational LSTM-RNN LM (line 25) on the ``clean", ``TF masking", ``Filter \& Sum" test conditions.
A statistically significant WER difference was only obtained by using the variational LSTM-RNN LM (line 25) over Bayesian/GP LSTM-RNN LMs (line 21, 23, 24) on the ``MVDR" beamforming test condition. 
When comparing the overall WER averaged over the four test conditions, again there was no statistically significant WER difference between the Bayesian/GP LSTM-RNN LMs (line 21, 22 and 24) and variational LSTM-RNN LM (line 25).}

4) The best performance was obtained using a 5-way interpolation between the baseline 4-gram, LSTM-RNN, Transformer LMs and their respective Gaussian Process counterparts (\revision{line} \revision{42}, last column) with their internal positions of applying GP modeling auto-configured using NAS. 
Statistically significant average WER reductions of {1.3\% and 1.2\% absolute (12.1\% and 11.3\% relative)} were obtained over the baseline LSTM-RNN (\revision{line} \revision{3}, last column) and Transformer LMs (\revision{line} \revision{4}, last column) respectively after model combination.

\vspace{-0.3em}

\section{Conclusion}
\label{sec:con}
This paper presents a full Bayesian learning framework including three methods to systematically account for the underlying uncertainty in state-of-the-art LSTM-RNN and Transformer LMs. 
The uncertainty over their model parameters, hidden activations and hidden output vectors are modeled using Bayesian, Gaussian Process and variational LSTM-RNN or Transformer LMs respectively. 
Novel inference approaches were used to speed up model inference and allow the computational cost in Bayesian NNLMs training and evaluation time comparable to those of the baseline LSTM-RNN and Transformer LMs with \revision{point estimate based} parameters. 
Experimental results obtained on the AMI meeting transcription and Oxford-BBC LipReading Sentences 2 overlapped speech recognition tasks suggest the proposed three Bayesian neural language modeling approaches can effectively mitigate the risk of over-fitting and poor generalization when LSTM-RNN and Transformer LMs with deterministic model parameters are trained on limited task domain data. 
\revision{Experimental results across multiple data sets and testing conditions suggest that the Bayesian or GP estimated LSTM-RNN or Transformer LMs with NAS auto-configured layer level uncertainty modeling and efficient inference using a minimal number of parameter samples can provide consistent performance improvements when being further linearly interpolated with the $n$-gram LM and respective baseline NNLMs using point estimation (e.g. line 24, 44 in Table \ref{table:3}, line 21, 32 in Table \ref{table:4}, and line 32, 34, 37 in Table \ref{table:6}).
Performance improvements from Bayesian learning can be retained when combining LSTM-RNN or Transformer LMs derived using standard or Bayesian estimation (e.g. line 42, Table \ref{table:6}).}
Future {research works} will investigate Bayesian learning approaches for fast domain adaptation of large scale pre-trained neural network language models. 
\vspace{-0.3em}

\section*{Acknowledgment}
This research is supported by Hong Kong Research Grants Council GRF grant No. 14200218, 14200220, 14200021, 
Innovation \& Technology Fund grant No. ITS/254/19 and InP/057/21. 

\vspace{-0.2em}


\bibliographystyle{IEEEtran}
\bibliography{main}

\end{document}